\documentclass{article} % For LaTeX2e
\usepackage{iclr2026_conference,times}
\iclrfinalcopy

\usepackage{amsmath,amsfonts,bm,amsthm}

\newtheorem{theorem}{Theorem}[section] % 定理按章节编号
\def\eqref#1{equation~\ref{#1}}
\def\1{\bm{1}}

\DeclareMathAlphabet{\mathsfit}{\encodingdefault}{\sfdefault}{m}{sl}
\SetMathAlphabet{\mathsfit}{bold}{\encodingdefault}{\sfdefault}{bx}{n}

\usepackage{hyperref}
\usepackage{url}
\usepackage{booktabs}

\usepackage{multirow}
\usepackage{makecell}
\usepackage{graphicx}
\usepackage{amsmath}
\usepackage{todonotes}
\usepackage{subcaption}
\usepackage{wrapfig}
\usepackage{pifont}
\usepackage[capitalize]{cleveref}
\crefname{section}{Section}{Sections}
\crefname{table}{Table}{Tables}
\crefname{figure}{Figure}{Figures}
\crefname{equation}{Equation}{Equations}

\newcommand{\methodabbr}{PM-KVQ}
\renewcommand{\cite}{\citep}

\title{\methodabbr: Progressive Mixed-precision KV Cache Quantization for Long-CoT LLMs}

\author{Tengxuan Liu\thanks{Equal contribution.} \ $^{1,2}$,
    Shiyao Li\thanks{Program leader.} \ $^{* 1,2}$,
    Jiayi Yang$^{* 3}$,
    Tianchen Zhao$^{1}$, 
    Feng Zhou$^{4}$, \\
    \textbf{Xiaohui Song$^{4}$,
    Guohao Dai$^{5,2}$, 
    Shengen Yan$^{2}$, 
    Huazhong Yang$^{1}$,
    Yu Wang\thanks{Corresponding author: Yu Wang (yu-wang@tsinghua.edu.cn).} \ $^{1}$} \\
    \\
  $^1$Tsinghua University, 
  $^2$Infinigence-AI, 
  $^3$Columbia University, \\
  $^4$OPPO AI Center, Beijing, China, 
  $^5$Shanghai Jiaotong University \\
}

\begin{document}

\maketitle

\begin{abstract}
Recently, significant progress has been made in developing reasoning-capable Large Language Models (LLMs) through long Chain-of-Thought (CoT) techniques.
However, this long-CoT reasoning process imposes substantial memory overhead due to the large Key-Value (KV) Cache memory overhead.
Post-training KV Cache quantization has emerged as a promising compression technique and has been extensively studied in short-context scenarios.
However, directly applying existing methods to long-CoT LLMs causes significant performance degradation due to the following two reasons: 
(1) \textbf{Large cumulative error}: Existing methods fail to adequately leverage available memory, and they directly quantize the KV Cache during each decoding step, leading to large cumulative quantization error.
(2) \textbf{Short-context calibration}: Due to Rotary Positional Embedding (RoPE), the use of short-context data during calibration fails to account for the distribution of less frequent channels in the Key Cache, resulting in performance loss.
We propose \textbf{P}rogressive \textbf{M}ixed-Precision \textbf{KV} Cache \textbf{Q}uantization (\textbf{\methodabbr{}}) for long-CoT LLMs to address the above issues in two folds:
(1) To reduce cumulative error, we design a progressive quantization strategy to gradually lower the bit-width of the KV Cache in each block. Then, we propose block-wise memory allocation to assign a higher bit-width to more sensitive transformer blocks. 
(2) To increase the calibration length without additional overhead, we propose a new calibration strategy with positional interpolation that leverages short calibration data with positional interpolation to approximate the data distribution of long-context data.
Extensive experiments on 7B–70B long-CoT LLMs show that \methodabbr{} improves reasoning benchmark performance by up to 8\% over SOTA baselines under the same memory budget and achieves 2.73–5.18$\times$ throughput over the original 16-bit LLMs.
Our code is available at \url{https://github.com/thu-nics/PM-KVQ}.
\end{abstract}

\section{Introduction}
\label{sec:intro}

% 研究的模型、数据、应用场景；引出现有范式的问题
Recently, many pioneers have developed remarkable reasoning Large Language Models (LLMs) with long Chain-of-Thoughts (CoT) techniques, such as OpenAI-o1~\citep{openai-o1}, DeepSeek-R1~\cite{deepseek-r1}, QwQ~\cite{qwq32b}, and so on.
To achieve better algorithmic performance, these long-CoT reasoning LLMs are trained to generate up to 128K tokens with multiple complex rationales from different perspectives~\cite{deepseek-r1}.
However, this long-CoT process demands significant memory overhead ($\sim$10GB–100GB) to store the Key-Value (KV) Cache as the history information, which limits the practical application scenarios for such long-CoT LLMs.

% 简要介绍现有的解决方法，说明现有解决方案的做法，并分析现有方法的困境，我们分析有a，b，c...几个关键难点问题
To mitigate the substantial memory overhead of long-CoT LLMs, various KV Cache compression methods have been proposed~\cite{kivi,mikv,rotatekv,streamingllm,moa}.
Among them, Post-training KV Cache Quantization is a promising compression technique that has already been well explored in short-context scenarios (e.g., $<$8K tokens).
% By applying post-training quantization, the memory footprint of the KV Cache can be significantly reduced.
QServe~\cite{qserve} and MiKV~\cite{mikv} observe that the Key Cache has more outliers than the Value Cache, leading to higher quantization error.
More importantly, the outliers in the Key Cache persist in certain channels.
To this end, they propose a channel-wise equalization method to migrate the outliers from the Key tensor to the Query tensor, thereby significantly reducing the quantization error.
KIVI~\cite{kivi}, SKVQ~\cite{skvq}, and IntactKV~\cite{intactkv} gain insights from the data distribution of the attention map and preserve the first or most recent tokens in higher bit-width within the KV Cache to maintain the performance.

However, directly applying the above short-context-optimized methods to long-CoT LLMs results in severe performance degradation.
The reasons stem from the following two aspects:
(1) \textbf{Large cumulative error in long-CoT LLMs}:
As a lossy compression method, directly quantizing the Key and Value tensors~\cite{kivi,qserve,mikv,skvq} introduces quantization errors at each decoding step when generating one token.
As the number of generated tokens increases, the accumulated quantization error grows larger, leading to a significant performance degradation of long-CoT LLMs.
(2) \textbf{Short calibration data cannot reflect long-context data distribution}:
The Rotary Positional Embedding (RoPE) operator incorporates positional information into each channel of the Key Cache by rotating token embeddings using sine and cosine functions of different frequencies.
For low-frequency channels after RoPE, which have a period of over 32K tokens, calibration using short sequences (e.g., 2K tokens) fails to accurately reflect the data distribution of the Key Cache, leading to more significant quantization errors.
% Due to the Rotary Positional Embedding (RoPE), the data in a Key Cache channel follows a sine-like pattern as the number of generated tokens increases, with the frequency related to the channel index.
% When calibration is performed using short sequences, channels with extremely low frequencies may not reach their peak values, resulting in an underestimation of preceding outlier magnitudes.
% \todo{@ltx, brief introduce and describe the effect of \textbf{RoPE}. Why short calibration will fail.}

% This issue primarily arises because these methods apply low-bit quantization to the KV Cache from the beginning of the inference process.
% Consequently, the available memory resources are not fully utilized, and the repeated use of imprecise KV Cache throughout the process leads to the accumulation of quantization errors over time.
% Moreover, current methods rely on short sequences during calibration to compute channel-wise reparameterization factors.
% When applied to long-CoT scenarios, these methods struggle to capture the characteristics of long-context KV Cache data distributions, often resulting in significant performance degradation.
% Although naively extending the calibration data to longer contexts can alleviate this issue, it incurs substantial increases in both time and memory consumption during calibration.

% 核心：为了解决这些难点问题，我们对应提出A，B，C几个解决方案。要用最凝练的话让大家体会到为什么我们的解决方案是有效的。（列出oracle实验，需要有数字/图表）
In this paper, we propose \textbf{P}rogressive \textbf{M}ixed-Precision \textbf{KV} Cache \textbf{Q}uantization (\textbf{\methodabbr{}}) to address the above two issues respectively.
(1) To reduce cumulative error, we aim to fully utilize the memory budget of the target hardware through two strategies.
On the one hand, we propose to quantize the KV Cache progressively.
For example, to achieve extremely low-bit quantization, such as 2-bit, instead of directly quantizing KV Cache to 2-bit at each decoding step, we initially store KV Cache in 16-bit format and then gradually reduce the bit-width to 2-bit through shifting operations once the memory resource is fully occupied.
On the other hand, we propose a block-wise memory allocation technique to allocate higher bit-widths for more sensitive blocks.
Specifically, we formalize the bit-width allocation task as an Integer Programming problem, which can be effectively solved by existing solvers with negligible latency.
% (1) To reduce cumulative error, we first allocate different memory budgets for KV Cache in different transformer blocks based on their sensitivity to quantization.
% Within this budget, we adopt a progressive quantization strategy by using high-bit-width quantization when the memory is sufficient and lower the bit-width when the allocated memory becomes saturated.
% \todo{@ltx, list the two insights to reduce the cumulative error here, block-wise and progressive. At least, introduce why we can do these optimizations.}
(2) To increase the effective calibration length without introducing additional computational or memory overhead, we retain the use of short-context data during calibration to maintain low resource consumption.
Furthermore, we propose leveraging positional interpolation~\cite{pi} to embed long-context positional information into short-context data, thereby enabling a more accurate estimation of the data distribution for long sequences.
% (2) To increase the calibration length without additional overhead, we leverage short calibration data with position index interpolation to approximate the data distribution of long-context data.
% \todo{@ltx, briefly describe our method.}

% \methodabbr{} begins by allocating a memory budget to each layer based on its sensitivity to quantization.
% During inference, \methodabbr{} employs a progressive quantization strategy, initially quantizing the KV Cache using a high bit-width.
% As the allocated memory becomes saturated, the bit-width of previously stored KV Cache entries is reduced to accommodate subsequent entries.
% By preserving high-precision KV Cache in the early stages of inference, \methodabbr{} maximizes the utilization of the memory budget and effectively mitigates the accumulation of quantization error throughout the process.
% Furthermore, we adopt an interpolation-based calibration method that leverages short-context data to approximate the data distribution encountered in long-CoT reasoning without incurring additional time or memory overhead during calibration.

% 总结本文的contribution
To sum up, the proposed \methodabbr{} mainly contains the following contributions:
\begin{itemize}
    \item We design progressive quantization and block-wise memory allocation techniques tailored for long-CoT scenarios to fully utilize the memory budget of the target hardware and effectively reduce cumulative quantization error.
    \item We propose to use short-context calibration data with positional interpolation to increase the effective length without incurring additional computational or memory overhead.
    \item Extensive experiments on long-CoT LLMs, ranging from 7B to 70B, show that the proposed \methodabbr{} achieves up to 8\% accuracy improvement over SOTA baselines on reasoning benchmarks under 4-bit/2-bit KV Cache quantization settings, while delivering a 2.73–5.18$\times$ throughput improvement over the 16-bit model. %\todo{@ltx, add eff results}
\end{itemize}

% （optional）可以在这里ref并总结一下每一个section是讲什么的。微缩版的目录

\section{Related Work}
\label{sec:related}

\subsection{Long CoT Large Language Models}
\label{sec:related-long}
Long-CoT (Long-Chain-of-Thought) LLMs aim to enhance multi-step reasoning capabilities for complex tasks like mathematical proofs, scientific reasoning, and multi-hop QA.
Models such as OpenAI-o1~\cite{openai-o1}, QwQ~\cite{qwq32b}, and DeepSeek-R1~\cite{deepseek-r1} employ advanced techniques to extend CoT reasoning depth.
DeepSeek, specifically, integrates iterative self-refinement and tool-augmented reasoning (e.g., code execution and symbolic solvers) to maintain coherence across extended reasoning chains.
Its architecture emphasizes hierarchical decomposition of problems and error-correction mechanisms, achieving state-of-the-art performance.

\begin{table}[htbp]
    \centering
    \caption{The memory overhead of the long-CoT LLMs. The batch size is 16, and the context length is 32K. }
    \begin{tabular}{c|c|c}
    \toprule
    Model & Weights (GB) & KV Cache (GB) \\
    \midrule
    DeepSeek-LLaMA-8B  & 16  & 64 \\ 
    DeepSeek-Qwen-32B   & 64  & 128 \\
    DeepSeek-LLaMA-70B & 140 & 160 \\
    \bottomrule
    \end{tabular}
    \label{tab:long-cot-llm-data}
\end{table}

While long-CoT can significantly improve model performance, it introduces excessively more decoding tokens (e.g., $>$32K tokens per request) and large GPU memory overhead. 
As shown in \cref{tab:long-cot-llm-data}, despite employing efficient attention designs, such as Multi-Query Attention (MQA)~\cite{mqa}, Group-Query Attention (GQA)~\cite{gqa}, and Multi-head Latent Attention (MLA)~\cite{mla}, the memory overhead of the KV Cache in long-CoT LLMs remains significantly large, often surpassing that of the model weights. 
Consequently, reducing the memory overhead of the KV Cache is  important for large batch sizes and long context requirements.

\subsection{Post-Training KV Cache Quantization}
\label{sec:related-quant}
To alleviate the large memory overhead with long reasoning contexts, many efforts have been made to reduce the KV Cache size.
Post-training KV Cache quantization stands as a promising technique for efficient inference.
KV Cache quantization methods try to use low bit-width integers to represent the cached key and value states, instead of using high bit-width floating-point values. 
Existing methods typically apply asymmetric uniform quantization for KV Cache:
\begin{equation}
\label{eq:integer_quant_main}
    \mathbf{X}_{\mathrm{asym}} = \left\lfloor\frac{\mathbf{X}_{\mathrm{BF16}}-Z}{S_{\mathrm{asym}}} \right\rceil,
\end{equation}
\begin{equation}
\label{eq:integer_quant_scale}
    S_{\mathrm{asym}} = \frac{\mathrm{max}(\mathbf{X}_{\mathrm{BF16}})-Z}{2^{b}-1},
\end{equation}
where $\mathbf{X}_{\mathrm{BF16}}$ denotes the 16-bit brain floating point (BF16) Key or Value tensor, $\mathbf{X}_{\mathrm{asym}}$ denotes the integer Key or Value tensor, $S_{\mathrm{asym}}$ and $Z=\min(\mathbf{X}_{\mathrm{BF16}})$ denote the scaling factor and the zero point respectively, $b$ denotes the quantization bit-width, $\lfloor\cdot\rceil$ denotes the rounding function. 

Specifically, MKLV~\cite{mklv} discovers that the sensitivity of Key and Value tensors are quite different, with the Key tensors being more sensitive to quantization than the Value tensors.
Therefore, MKLV simply assigns a higher bit-width to Key tensors and a lower bit-width to Value tensors.
WKVQuant~\cite{wkvquant} proposes to change the data flow of the previous KV Cache quantization by using the unquantized current Key and Value to calculate the attention operator, and then quantize the current Key and Value.
% By doing so, WKVQuant can significantly reduce the cumulative quantization error for the current token.
SKVQ~\cite{skvq} further improves the WKVQuant by using a sliding window that stores the most recent 128 Key and Value features in floating-point format to reduce the cumulative quantization error.
MiKV~\cite{mikv} is inspired by H2O~\cite{h2o} to use the heavy-hitter oracle to discover the important tokens in a higher bit-width and quantize the rest of the unimportant tokens into a lower bit-width.
KIVI~\cite{kivi} discovers that the Value tensors are much flatter than Key tensors, and the outliers in Key tensors typically appear in certain channels. 
To this end, KIVI utilizes per-channel quantization for Key Cache and per-token quantization for Value Cache in a group-wise manner to reduce the quantization error.
RotateKV~\cite{rotatekv} combines the channel-wise equalization and the rotation-based equalization with Hadamard matrices to further reduce the quantization error.

% In this paper, we adopt effective strategies from prior work, consistently storing the first token in INT16 format and using a sliding window to preserve the most recent few tokens.
% To further reduce quantization errors, we introduce two novel improvements:
% (1) Progressive Quantization: We maximize memory utilization by initially storing KV Cache in higher precision. When the memory is full, we progressively reduce the bit-width to store more Keys and Values.
% (2) Block-wise Memory Allocation: When memory is large enough to store more tokens with a higher bit-width, we will allocate more memory to those sensitive transformer blocks to preserve the performance.

In this paper, we adopt effective strategies from prior work, such as storing the first token in INT16 and using a sliding window for recent tokens. To further reduce quantization errors, we propose two improvements:
(1) Progressive Quantization – initially store KV cache in higher precision and gradually lower the bit-width as memory memory becomes saturated;
(2) Block-wise Memory Allocation – allocate more memory to sensitive transformer blocks when capacity allows, thereby preserving performance.

% To tackle the outliers present in the KV Cache, these methods extract the outlier as full precision or use
% finer-grained quantization scheme, which increases the quantization overhead. In this study,
% we propose an efficient channel-separable quantization scheme with reduced quantization overhead
% and strong performance. Additionally, both categories of methods commonly adopt accumulated
% attention scores as the metric for token importance [46, 43]. However, we observe that this criterion
% is inaccurate and can result in significant performance deterioration at low bit-widths. In contrast, we
% achieve superior compression performance by utilizing a more accurate metric for identifying salient
% tokens.

\section{Method}
\label{sec:method}

\begin{figure}[t]
    \centering
    % \vspace{-10pt}
    \includegraphics[width=\linewidth]{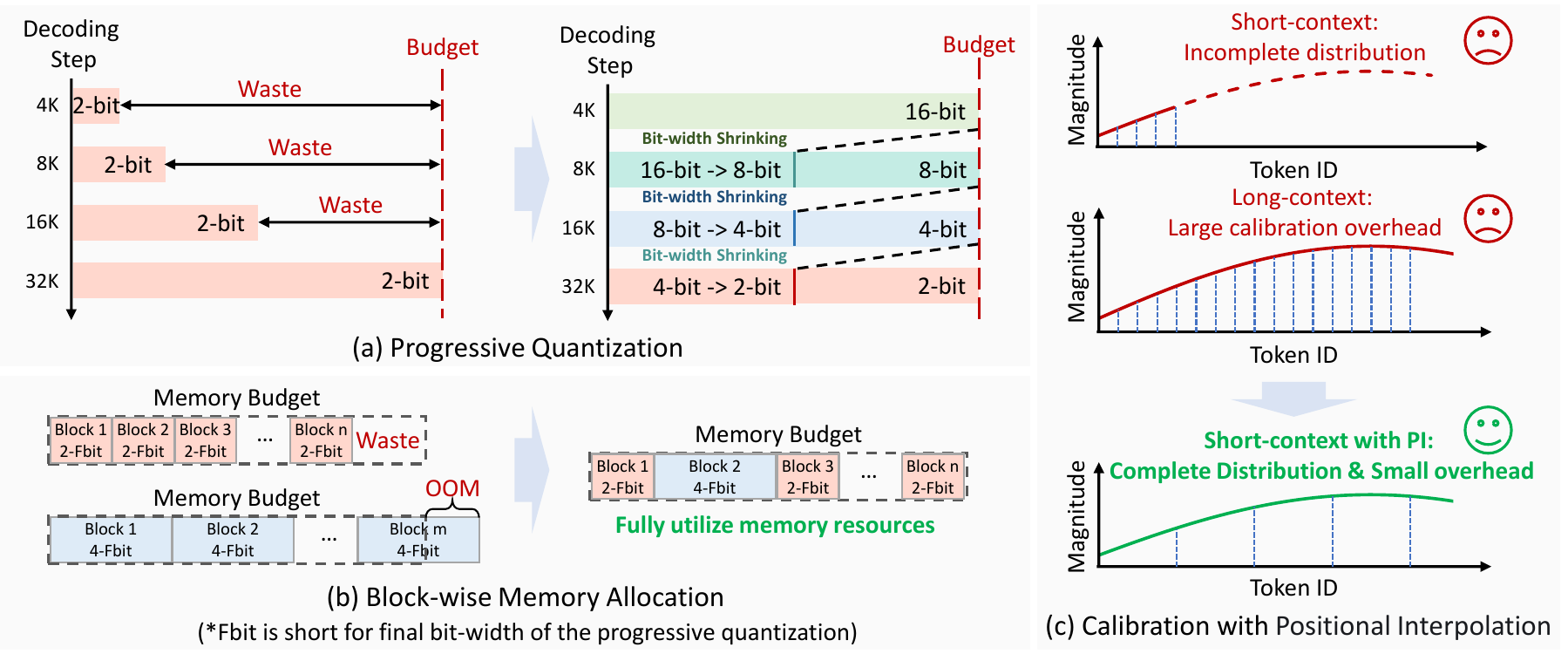}
    \vspace{-20pt}
    \caption{Method Overview. (a) Progressive quantization: we progressively shrink the bit-width of KV Cache to fully utilize the memory budget. (b) Block-wise memory allocation: we allocate a higher bit-width to those transformer blocks with higher sensitivity. (c) Calibration with Positional Interpolation to approximate the distribution of long-context data with short-context data.}
    \label{fig:overview}
    \vspace{-10pt}
\end{figure}

\vspace{-2pt}
\subsection{Progressive Quantization}
\label{sec:method-token}
\vspace{-1pt}
As discussed in \cref{sec:related-quant}, existing post-training KV Cache quantization methods quantize at every decoding step, causing large cumulative errors.
A sliding window with high-precision cache alleviates this, but very low bit-widths (e.g., 2-bit) still lead to severe accuracy loss in long-CoT tasks.
% Besides, existing methods waste a lot of memory at the initial stage of the inference process. 这句话是错的，什么事initial stage？在4bit及之前都浪费了吧？
% We demonstrate that existing KV Cache quantization approaches underutilize the allocated memory budget, missing critical opportunities to reduce cumulative quantization errors.
\textbf{We show that existing KV cache quantization methods underutilize the memory budget and miss opportunities to reduce cumulative errors.}
% As shown in \cref{fig:overview}(a) left, the output sequence is extremely long (e.g., 32K) in the long-CoT scenario, and the hardware is capable to accommodating the KV Cache for such a sequence length. 这句话写出来也太扯了。你凭什么能保证hardware一定能分配足够的显存的，我总共给你8G的显存，连模型都放不下。这里其实就举个例子说明存在性就行了，这种情况一定是存在的，话说的太满必然会被challenge。
% However, in the early decoding steps when the output sequence is relatively short, a considerable portion of the memory remains underutilized. 还是同一个问题，概念不清楚，什么是early decoding step？从来没定义过
% Existing methods quantize the KV Cache to a low bit-width at this time, causing a waste of memory resources. “at this time” 指的是什么time？ 
% For example, when we use SOTA methods to achieve 2-bit KV Cache quantization, the resulting memory consumption of the KV Cache is illustrated in the left panel of \cref{fig:overview}(a).
% As we can see, existing SOTA methods store 2-bit KV Cache at every decoding step, leading to significant memory waste when the memory budget is not fully utilized.
As illustrated in the left panel of \cref{fig:overview}(a), SOTA methods store 2-bit KV Cache at every decoding step, causing substantial memory waste when the budget is not fully used.

% To address these issues, we propose a progressive quantization strategy aimed at reducing the cumulative quantization error at each decoding step and maximizing the utilization of hardware memory resources. 上面是几个issue？哪里来的issues？逻辑不清晰，降低累积误差和充分利用存储资源是并列关系吗？并列关系的话那就是我可以只解决一个不管另一个了，那这两个事情的性质也都不同啊，前者是最终目的，后者是动机。
% Progressive quantization involves initially preserving a high-precision KV Cache and gradually reducing the bit-width of the KV Cache during the long-context decoding process.
To address the above issue, we propose a progressive quantization strategy to make full use of the memory resources by gradually shrinking the bit-width of the KV Cache, thereby significantly reducing the cumulative quantization error.
% Specifically, for each transformer block, we predefine the final bit-width of the progressive quantization as ``Fbit'' and, based on this value, calculate the memory budget for KV Cache. 
\textit{For each transformer block, we use ``Fbit'' to represent the final bit-width of the progressive quantization process.}
In this case, we can easily calculate the memory budget for each transformer block based on the maximum context length of the target long-CoT LLM.
As shown in \cref{fig:overview}(a) right, the Fbit in this example is 2-bit and the maximum context length is 32K. 
During generation, we initially store the KV Cache in 16-bit format to alleviate the large cumulative quantization error. 
\textbf{Once the memory budget is fully utilized}, we apply a bit-width shrinking strategy to accommodate more tokens by progressively reducing the bit-width of the existing KV Cache.
Specifically, we use powers of two for quantization bit-widths, gradually decreasing them in the order of 16, 8, 4, and 2 bits.

In addition, for the bit-width shrinking strategy, we design an ``\textbf{Equivalent Right Shift}'' strategy that is mathematically equivalent to de-quantizing the $2b$-bit KV Cache and then quantizing it to $b$-bit.
Here, $b$ can be 8, 4, or 2, corresponding to shrinking the KV Cache from 16-bit to 8-bit, 8-bit to 4-bit, and 4-bit to 2-bit, respectively.
Specifically, we formulize the bit-width shrinking strategy by using integer addition and shifting as follows:
\begin{equation}
    \mathbf{X}_{b}=\left((2^{2b}-2^b+1)(\mathbf{X}_{2b}+2^{b-1})\right) >> 3b,
\label{eq:equivalent_shift}
\end{equation}
where $\mathbf{X}_{b}$ and $\mathbf{X}_{2b}$ represent the $b$-bit and $2b$-bit tensor respectively.
We keep the zero point unchanged ($Z_b = Z_{2b}$) and increase the scaling factor to $S_b = (2^b + 1) S_{2b}$ to preserve the dynamic range of the data distribution.
The detailed proof of equivalence for \cref{eq:equivalent_shift} is shown in \cref{add:proof}.
Furthermore, we compare the effect of three different bit-width shrinking strategies and show that the ``Equivalent Right Shift'' strategy achieves better performance, as detailed in \cref{sec:exp-ablation-pm}.

\vspace{-2pt}
\subsection{Block-wise Memory Allocation}
\label{sec:method-layer}
% Existing KV Cache quantization methods typically apply a uniform bit-width across all transformer blocks.
% However, uniform memory allocation may lead to a waste of the memory budget.
Existing KV Cache quantization methods typically apply a uniform bit-width across all transformer blocks, which may not fully utilize the memory resources of the target hardwares.
% For example, in the circumstance of \cref{fig:overview}(b) left, uniform 2-Fbit KV Cache may cause a memory waste while uniform 4-Fbit KV Cache may cause an out-of-memory error. 
As shown in \cref{fig:overview}(b) left, in this example, the target hardware has sufficient memory to store the KV Cache uniformly in 2-Fbit format, leaving a proportion of wasted memory. 
However, switching to a uniform 4-Fbit format may exceed the memory limit and trigger an out-of-memory error.
Therefore, using a uniform bit-width for KV Cache may not fully utilize the available memory across different scenarios with varying memory resources.

% To set Fbit for each transformer block while fully utilize the memory budget, we adopt a transformer-block-wise memory allocation strategy. 这句话也存在严重的逻辑问题，每一个词组的关系是什么？set Fbit是方法的具体说明，fully utilize是目的。
To fully utilize the memory resource in different scenarios for better performance, we propose a block-wise memory allocation strategy to assign a higher bit-width for more sensitive blocks.
% Inspired by the metric used in mixed-precision weight quantization~\cite{llm-mq,mixdq}, we first assess the sensitivity to quantization of the KV Cache in each block using the first-order Taylor approximation:
Inspired by existing mixed-precision quantization methods~\cite{llm-mq,mixdq}, we employ a first-order Taylor approximation to estimate the sensitivity of the model output to perturbations in the Key Cache and Value Cache.
Here, we take the Key Cache as an example:
\begin{equation}
    \mathcal{L}(Q_b(\mathbf{K}_i)) \approx \mathcal{L}(\mathbf{K}) + \mathbf{G}_{\mathbf{K}_i}\odot(\mathbf{K}_i-Q_b(\mathbf{K}_i)),
\end{equation}
where $\mathcal{L}$ is the loss function, $i$ represents the $i$-th transformer block, $\mathbf{K}_i$ is the Key Cache, $Q_b(\cdot)$ is the $b$-bit quantization function, $\mathbf{G}_{\mathbf{K}_i}$ is the gradients of the loss function with respect to the $\mathbf{K}_i$, $\odot$ is the element-wise multiplication operator.
The Value Cache follows a similar formulation.
% sensitivity to quantization of the KV Cache in each block using the first-order Taylor approximation: 你这里这个公式根本就不是一阶泰勒近似

To minimize the effect of KV Cache quantization in each transformer block, we aim to minimize the following sensitivity term:
\begin{equation}
    s_{i,b}=\|\mathbf{G}_{\mathbf{K}_i}\odot(\mathbf{K}_i-Q_b(\mathbf{K}_i))\|_1+\|\mathbf{G}_{\mathbf{V}_i}\odot(\mathbf{V}_i-Q_b(\mathbf{V}_i))\|_1,
\label{eq:sensitivity}
\end{equation}
where $s_{i,b}$ denotes the sensitivity of the KV Cache in the $i$-th transformer block to $b$-bit quantization.

% As illustrated in Figure X, the sensitivity of KV Cache quantization varies substantially across different transformer blocks.
% Oracle experiments further demonstrate that increasing the bit-width for the most sensitive KV Cache leads to significant performance gains.

% To derive the optimal memory allocation strategy, we formulate the memory allocation as the following Integer Programming Problem: 这句话干巴巴地放在这里和前面的逻辑没有连起来呀，为什么要列这个式子？和前面的s_{i,b}有什么关系？看了只会懵逼
Taking into account the sensitivity of all transformer blocks, our goal is to assign an appropriate bit-width to each block to minimize the impact on the loss function, subject to a given memory budget.
To this end, we formulate the block-wise bit-width allocation as the following Integer Programming problem:
\begin{align}
    \mathop{\arg \min}_{x_{i,b}} &\enspace \sum_i^N \sum_{b} x_{i,b} \cdot s_{i,b}, \\
    & \displaystyle\sum_b x_{i,b}=1,
    \displaystyle\sum_i^N \sum_b x_{i,b} \cdot \left( Mem(Q_{b}(\mathbf{K}_{i}))+Mem(Q_{b}(\mathbf{V}_{i})) \right) \leq \mathcal{M}, \\
    & x_{i,b} \in \left\{0,1\right\}, b \in B,
\label{eq:allocation}
\end{align}
where $N$ is the number of transformer blocks, $Mem(\cdot)$ is the function to calculate the memory usage of the quantized KV Cache, $\mathcal{M}$ is the memory budget for the KV Cache of all the transformer blocks, $x_{i,b}$ is the one-hot vector that indicates the bit-width choice $b$ of the $i$-th block, and $B$ is the optional bit-width set, detailed in \cref{sec:exp-setups-budget}.
The proposed Integer Programming problem can be effectively solved by CVXPY~\cite{cvxpy} within a few seconds.
% By solving this optimization problem using an efficient solver, we can make full use of the memory budget and reduce the cumulative quantization error.

\subsection{Calibration with Positional Interpolation}
\label{sec:method-calib}
Previous studies have observed that the Key Cache of LLMs contains outliers in certain channels, which significantly increases the quantization error.
Approaches such as QServe~\cite{qserve} address this issue by introducing a channel-wise reparameterization method to transfer the outliers in Key tensors into Query tensors:
\begin{equation}
    \mathbf{P}=(\mathbf{Q\Lambda})\cdot Q\left((\mathbf{K\Lambda}^{-1})^T\right), \mathbf{\Lambda}=diag(\lambda_i),
\label{eq:rep}
\end{equation}
where $i$ is the channel index, $\lambda_i$ is the reparameterization factor of the $i$-th channel, and $Q(\cdot)$ is the quantization function.
Generally, $\lambda_i$ is calibrated using a small dataset of sequences with a typical length of 512 tokens, which is much shorter than the maximum output length of 32K tokens.
The calibration process follows \cref{eq:calibration}:
\begin{equation}
    \lambda_i={\left(\max_m K_{m,i}\right)}^{\alpha},
\label{eq:calibration}
\end{equation}
where $m$ is the token position index, and $\alpha$ is the parameter to adjust the strength of outlier transfer, which can be set as a fixed number or obtained by grid search~\cite{awq}.

However, applying the above reparameterization technique to long-CoT LLMs using short calibration data (e.g., 512) may fail to accurately capture the distribution of the Key Cache.
This limitation arises because Rotary Positional Embedding (RoPE)~\cite{rope} is used to inject positional information into the Key Cache, which introduces periodic variations across different channels:
\begin{equation}
    \begin{bmatrix}
        \widetilde{K}_{m,i} \\ 
        \widetilde{K}_{m,i+\frac{d}{2}}
    \end{bmatrix}=
    \begin{bmatrix}
        \cos m\theta_{i} & -\sin m\theta_{i} \\
        \sin m\theta_{i} & \cos m\theta_{i}
    \end{bmatrix}
    \begin{bmatrix}
        K_{m,i} \\
        K_{m,i+\frac{d}{2}}
    \end{bmatrix}=
    \sqrt{K_{m,i}^2+K_{m,i+\frac{d}{2}}^2}
    \begin{bmatrix}
        \cos(m\theta_{i}+\varphi) \\
        \sin(m\theta_{i}+\varphi)
    \end{bmatrix},
\label{eq:rope}
\end{equation}
where $K$ and $\widetilde{K}$ denote the Keys before and after RoPE respectively, $d$ is the hidden dimension of each attention head, and $\theta_{i}$ denotes the rotary frequency of channel $i$ and $i+d/2$.
Since $\theta_{i}=\theta^{-2i/d}$ decreases with increasing $i$, the frequency of the sine curve is extremely low in channels with indices near $d/2$ and $d$.
For example, in the DeepSeek-R1-Distill-Qwen-7B, the lowest frequency sine curve has a period of up to 54,410 tokens.
Therefore, when using short sequences of 512 tokens for calibration, as shown in \cref{fig:overview}(c) top, we cannot obtain the reparameterization factor that can completely reflect the sine-like data distribution.

% Directly increasing the calibration sequence length significantly increases both latency and memory costs due to the $O(N^2)$ complexity of the self-attention mechanism. (1)什么叫latency cost？latency和memory是对称的说法吗？latency指的是时间结果、memory指的是硬件指标。(2)哪里提到过calibration sequence？不要重复定义概念 （3）mechanism是一个抽象的概念，不如过不是改变算法的流程那就用operator
Directly increasing the length of calibration data significantly increases both latency and memory costs due to the $O(N^2)$ complexity of the self-attention operator.
Instead, we embed long-context positional information into short calibration data by leveraging positional interpolation~\cite{pi}.
Specifically, we multiply a position scaling factor $s$ to the position index $m$ in the rotary matrix of RoPE for positional interpolation, as shown below: 
{ \small
\begin{equation}
    \begin{aligned}
    &\begin{bmatrix}
        \widetilde{K}_{m,i} \\ 
        \widetilde{K}_{m,i+\frac{d}{2}}
    \end{bmatrix}=
    \begin{bmatrix}
        \cos (s \cdot m\theta_{i}) & -\sin (s \cdot m\theta_{i}) \\
        \sin (s \cdot m\theta_{i}) & \cos (s \cdot m\theta_{i})
    \end{bmatrix}
    \begin{bmatrix}
        K_{m,i} \\
        K_{m,i+\frac{d}{2}}
    \end{bmatrix}
    = \sqrt{K_{m,i}^2+K_{m,i+\frac{d}{2}}^2}
    \begin{bmatrix}
        \cos(s \cdot m\theta_{i}+\varphi) \\
        \sin(s \cdot m\theta_{i}+\varphi)
    \end{bmatrix}.
    \end{aligned}
    \label{eq:interpolation-rope}
\end{equation}
}

As shown in \cref{fig:overview}(c) bottom, by applying positional interpolation, we can increase the largest positional index by $s\times$ without additional computation and memory overhead.

\subsection{Method Pipeline}
\label{sec:method-overview}
In this paper, the proposed \methodabbr{} combines the above three techniques to achieve better long-CoT performance with low bit-width KV Cache quantization.
(1) Before the inference process, we first profile the sensitivity of each transformer block based on the calibration dataset, detailed in \cref{sec:exp-datasets}, and solve the Integer Programming problem to set the proper Fbit for each transformer block, as discussed in \cref{sec:method-layer}.
Then, we apply the channel-wise reparameterization technique by using the calibration dataset with positional interpolation, as detailed in \cref{sec:method-calib}.
(2) During the inference process, we apply progressive quantization to the KV Cache by gradually lowering the bit-width from 16-bit to the allocated Fbit, as shown in \cref{sec:method-token}.
\section{Experiments}
\label{sec:exp}

\subsection{Experimental Setups}
\label{sec:exp-setups}

\subsubsection{Datasets}
\label{sec:exp-datasets}
\textbf{For the calibration dataset}, we use the arXiv subset of RedPajama~\cite{redpajama} as calibration dataset.
This subset consists of academic papers, containing mathematical formulas and reasoning process.
We randomly select 512 samples, each with a length of 2,048 tokens, for calibration.
For positional interpolation, we set $s=4$ in \cref{eq:interpolation-rope}, which means we embed an 8,192 context length to 2,048 tokens.
We set $\alpha$ in \cref{eq:calibration} by grid searching over [0,1] for the optimal $\alpha$ that minimizes the reconstruction loss of the self-attention operator with a grid size of 20.

\textbf{For performance evaluation}, we mainly focus on evaluating the long-CoT LLMs on the mathematical reasoning and code generation benchmarks with \textbf{long generation contexts ($>$16K)}.
For mathematical reasoning, we use the AIME-2024/2025~\cite{aime} and CMIMC-2024~\cite{cmimc} datasets.
For competition-level code generation, we select coding problems released between January 1, 2025, and April 6, 2025, from LiveCodeBench~\cite{livecodebench}.
% We also evaluate the instruction following capability of the LLMs using the IFEval~\cite{ifeval} dataset, as discussed in \cref{add:non-reasoning}.
Besides, as illustrated in \cref{add:non-reasoning}, we also evaluate the proposed \methodabbr{} on the IFEval~\cite{ifeval} dataset with \textbf{short generation contexts ($\sim$1K)} to demonstrate its strong generalizability across different context lengths.
We sample 16 responses for each mathematical problem, 4 responses for each code generation problem, and 1 response for each instruction following problem, using a temperature of 0.6, top-p of 0.95, and a maximum output length of 32,768 tokens.

\subsubsection{Baselines and Model Choice}
\label{sec:exp-setups-model}
\textbf{For baselines}, we compare \methodabbr{} with SOTA KV Cache quantization methods, including the uniform bit-width methods RotateKV~\cite{rotatekv}, KIVI~\cite{kivi}, and mixed-precision quantization method MiKV~\cite{mikv}, which retains the KV Cache of heavy hitters in BF16 format and uses low bit-width for other tokens.
We also compare \methodabbr{} with KVTuner~\cite{kvtuner} in \cref{add:kvtuner}.
Similar to KIVI, \methodabbr{} stores the KV Cache for the first and most recent 128 tokens in INT16 format to mitigate performance degradation.
All model weights in our experiments are in BF16 format.

\textbf{For model choices}, we evaluate the different quantization methods above on the Deepseek-R1-Distill~\cite{deepseek-r1} series as well as the QwQ-32B model~\cite{qwq32b}.
Specifically, the Deepseek-R1-Distill series is an LLM family distilled from DeepSeek-R1.
We choose Deepseek-R1-Distill-Qwen-7B/14B/32B and Deepseek-R1-Distill-LLaMA-8B/70B, ranging from 7B to 70B.

\vspace{-2pt}
\subsubsection{Bit-width and Batch Size Setups}
\label{sec:exp-setups-budget}
\textbf{For the bit-width settings}, to demonstrate the effectiveness of the proposed \methodabbr{}, we select quantization bit-widths that lead to significant performance degradation when using baseline methods for each long-CoT LLM. 
Specifically, we use 4-bit for DeepSeek-LLaMA-8B and 2-bit for other LLMs.
Notably, the bit-width for the proposed \methodabbr{} stands for the Fbit, as discussed in \cref{sec:method-token}.
In addition, for the optional bit-width set $B$ in \cref{sec:method-layer}, we use $B = \{4, 8\}$ for DeepSeek-LLaMA-8B, and $B = \{2,4\}$ for other long-CoT LLMs.
We use asymmetric group-wise quantization for KV Cache with a group size of 128, as shown in \cref{eq:integer_quant_main}.
All of the performance results are conducted with fake quantization on an 8$\times$A100-80G GPU server.

\textbf{For the batch size setups}, we assign a target GPU with different memory resources for different LLMs to show the memory constraints in real-world scenarios, as shown in \cref{tab:main}. 
On the one hand, to demonstrate the effectiveness of progressive quantization, we set the batch size for each LLM such that all methods can fully utilize the memory resources of the target GPU.
Specifically, we use a batch size of 8 for LLaMA-8B with a 4-bit KV Cache, 40 for Qwen-7B with a 2-bit KV Cache, and 16 for the other LLMs, as shown in \cref{tab:main}.
On the other hand, to evaluate the effectiveness of block-wise memory allocation, we use smaller batch sizes to allocate more memory per instance, ensuring that higher bit-widths cannot be directly used under the same constraints.
In this setting, we use a batch size of 6 for LLaMA-8B with a 4-bit KV Cache, 32 for Qwen-7B with a 2-bit KV Cache, and 12 for the remaining LLMs, as also shown in \cref{tab:main}.
Results across more target hardwares can be found in \cref{add:hardware}.
% We assign a target GPU and two batch sizes to simulate different memory budgets for the KV Cache. 这句话单独出现毫无信息量，你怎么选的batch？为什么这么选？你选bs=2也是不同的memory budget，你意识到了吗？
% For each setting, the quantization bit-width of the RotateKV and KIVI methods is configured to be the largest power-of-two integer that accommodates the KV Cache within the memory budget.
% The maximum number of heavy hitters is calculated based on the memory budget.

\begin{table}
  \caption{Main results of long-CoT Language Models on reasoning-related benchmarks with SOTA KV Cache quantization methods. ``BS'' is short for ``batch size''.}
  % \vspace{-5pt}
  \label{tab:main}
  \centering
  \resizebox{\linewidth}{!}{
  \begin{tabular}{llccccccccc}
    \toprule
    Models & Quantization & Bit-width & \multicolumn{2}{c}{AIME-2024} & \multicolumn{2}{c}{AIME-2025} & \multicolumn{2}{c}{CMIMC-2024} & LiveCode \\
    \cmidrule(lr){3-3}\cmidrule(lr){4-5}\cmidrule(lr){6-7}\cmidrule(lr){8-9}\cmidrule(lr){10-10}
    (Target GPU) & Methods & (K-V) & pass@1 & Voting & pass@1 & Voting & pass@1 & Voting & pass@1 \\

    \midrule
    
    ~ & - - & 16-16 & 41.04\scriptsize{$\pm$6.74} & 63.33 & 30.00\scriptsize{$\pm$3.33} & 36.67 & 27.29\scriptsize{$\pm$5.17} & 43.33 & 26.29\scriptsize{$\pm$1.34} \\
    ~ & RotateKV (BS=32,40) & 2-2 & 0.00\scriptsize{$\pm$0.00} & 0.00 & 0.00\scriptsize{$\pm$0.00} & 0.00 & 0.00\scriptsize{$\pm$0.00} & 0.00 & 0.00\scriptsize{$\pm$0.00} \\
    DeepSeek- & MiKV (BS=32) & 2/16-2/16 & 0.00\scriptsize{$\pm$0.00} & 0.00 & 0.63\scriptsize{$\pm$0.02} & 3.33 & 2.29\scriptsize{$\pm$0.02} & 3.33 & 5.86\scriptsize{$\pm$0.85} \\
    Qwen-7B & MiKV (BS=40) & 2-2 & 0.00\scriptsize{$\pm$0.00} & 0.00 & 0.00\scriptsize{$\pm$0.00} & 0.00 & 0.00\scriptsize{$\pm$0.00} & 0.00 & 0.00\scriptsize{$\pm$0.00}  \\
    (1$\times$4090-24G) & KIVI (BS=32,40) & 2-2 & 32.08\scriptsize{$\pm$5.25} & 43.33 & 24.58\scriptsize{$\pm$3.51} & 33.33 & 20.83\scriptsize{$\pm$3.63} & 23.33 & 19.00\scriptsize{$\pm$2.37} \\
    \cmidrule(lr){2-10}
    ~ & \methodabbr{} (BS=32) & 2/4-2/4 & \textbf{40.21}\scriptsize{$\pm$5.71} & \textbf{66.67} & \textbf{28.96}\scriptsize{$\pm$4.20} & \textbf{40.00} & 25.83\scriptsize{$\pm$5.20} & \textbf{40.00} & \textbf{24.71}\scriptsize{$\pm$1.48} \\
    ~ & \methodabbr{} (BS=40) & 2-2 & 40.00\scriptsize{$\pm$5.40} & 60.00 & 28.12\scriptsize{$\pm$4.71} & 33.33 & \textbf{26.46}\scriptsize{$\pm$4.64} & \textbf{40.00} & 24.57\scriptsize{$\pm$1.42} \\
    
    \midrule
    
    ~ & - - & 16-16 & 44.17\scriptsize{$\pm$4.49} & 66.67 & 30.63\scriptsize{$\pm$6.58} & 50.00 & 26.67\scriptsize{$\pm$4.41} & 36.67 & 32.14\scriptsize{$\pm$1.99} \\
    ~ & RotateKV (BS=6,8) & 4-4 & 42.92\scriptsize{$\pm$3.89} & 66.67 & 27.29\scriptsize{$\pm$6.48} & 40.00 & 26.46\scriptsize{$\pm$5.33}
    & 30.00 & \textbf{32.00}\scriptsize{$\pm$1.56} \\
    DeepSeek- & MiKV (BS=6) & 4/16-4/16 & 35.63\scriptsize{$\pm$7.14} & 66.67 & 24.79\scriptsize{$\pm$3.72} & 36.67 & 25.21\scriptsize{$\pm$3.53} & 33.33 & 27.00\scriptsize{$\pm$1.30} \\
    LLaMA-8B & MiKV (BS=8) & 4-4 & 41.67\scriptsize{$\pm$6.56} & 60.00 & 26.46\scriptsize{$\pm$7.02} & 43.33 & 22.92\scriptsize{$\pm$4.84} & 26.67 & 29.71\scriptsize{$\pm$1.67} \\
    (1$\times$4090-24G) & KIVI (BS=6,8) & 4-4 & 41.25\scriptsize{$\pm$6.65} & 60.00 & 27.92\scriptsize{$\pm$4.70} & 46.67 & 26.25\scriptsize{$\pm$4.98} & 36.67 & 30.29\scriptsize{$\pm$1.76} \\
    \cmidrule(lr){2-10}
    ~ & \methodabbr{} (BS=6) & 4/8-4/8 & \textbf{47.71}\scriptsize{$\pm$6.84} & \textbf{73.33} & \textbf{31.25}\scriptsize{$\pm$5.64} & \textbf{50.00} & 28.13\scriptsize{$\pm$4.08} & 36.67 & 31.71\scriptsize{$\pm$0.86} \\
    ~ & \methodabbr{} (BS=8) & 4-4 & 43.33\scriptsize{$\pm$5.57} & 63.33 & \textbf{31.25}\scriptsize{$\pm$5.64} & \textbf{50.00} & \textbf{28.96}\scriptsize{$\pm$5.10} & \textbf{40.00} & 31.57\scriptsize{$\pm$1.17} \\
    
    \midrule
    
    \multirow{4}{*}{\makecell[l]{DeepSeek-\\Qwen-14B\\(1$\times$A100-40G)}} & - - & 16-16 & 68.13\scriptsize{$\pm$7.26} & 80.00 & 50.00\scriptsize{$\pm$5.77} & 60.00 & 49.58\scriptsize{$\pm$4.84} & 66.67 & 45.71\scriptsize{$\pm$1.34} \\
    ~ & KIVI (BS=12,16) & 2-2 & 48.13\scriptsize{$\pm$4.85} & 70.00 & 33.96\scriptsize{$\pm$3.17} & 43.33 & 27.71\scriptsize{$\pm$3.67} & 33.33 & 34.43\scriptsize{$\pm$3.11} \\
    \cmidrule(lr){2-10}
    ~ & \methodabbr{} (BS=12) & 2/4-2/4 & \textbf{67.71}\scriptsize{$\pm$6.94} & 80.00 & \textbf{46.67}\scriptsize{$\pm$7.36} & \textbf{60.00} & \textbf{47.71}\scriptsize{$\pm$4.20} & 60.00 & \textbf{42.14}\scriptsize{$\pm$0.95} \\
    ~ & \methodabbr{} (BS=16) & 2-2 & 63.33\scriptsize{$\pm$4.08} & \textbf{83.33} & 42.08\scriptsize{$\pm$6.55} & \textbf{60.00} & 46.67 \scriptsize{$\pm$5.27}& \textbf{70.00} & 41.86\scriptsize{$\pm$1.78} \\
    
    \midrule
    
    \multirow{4}{*}{\makecell[l]{DeepSeek-\\Qwen-32B\\(1$\times$A100-80G)}} & - - & 16-16 & 72.08\scriptsize{$\pm$4.39} & 86.67 & 53.12\scriptsize{$\pm$5.71} & 66.67 & 52.50\scriptsize{$\pm$5.71} & 70.00 & 46.86\scriptsize{$\pm$2.18} \\
    ~ & KIVI (BS=12,16) & 2-2 & 63.96\scriptsize{$\pm$6.89} & \textbf{83.33} & 45.42\scriptsize{$\pm$5.38} & 60.00 & 40.63\scriptsize{$\pm$5.17} & 56.67 & 40.43\scriptsize{$\pm$1.10} \\
    \cmidrule(lr){2-10}
    ~ & \methodabbr{} (BS=12) & 2/4-2/4 & \textbf{69.17}\scriptsize{$\pm$5.95} & \textbf{83.33} & 48.54\scriptsize{$\pm$5.89} & 60.00 & \textbf{51.25}\scriptsize{$\pm$4.70} & 66.67 & \textbf{43.57}\scriptsize{$\pm$1.64} \\
    ~ & \methodabbr{} (BS=16) & 2-2 & 67.29\scriptsize{$\pm$4.89} & \textbf{83.33} & \textbf{48.96}\scriptsize{$\pm$7.33} & \textbf{63.33} & 50.42\scriptsize{$\pm$7.16} & \textbf{73.33} & \textbf{43.57}\scriptsize{$\pm$0.62} \\

    \midrule
    
    ~ & - - & 16-16 & 78.54\scriptsize{$\pm$4.85} & 86.67 & 67.71\scriptsize{$\pm$3.48} & 76.67 & 71.25\scriptsize{$\pm$3.51} & 80.00 & 54.71\scriptsize{$\pm$0.74} \\
    QwQ-32B & KIVI (BS=12,16) & 2-2 & 61.25\scriptsize{$\pm$5.51} & 76.67 & \textbf{51.67}\scriptsize{$\pm$5.27} & 63.33 & 48.33\scriptsize{$\pm$5.77} & 63.33 & 41.86\scriptsize{$\pm$1.21} \\
    \cmidrule(lr){2-10}
    (1$\times$A100-80G) & \methodabbr{} (BS=12) & 2/4-2/4 & 66.46\scriptsize{$\pm$3.81} & \textbf{80.00} & 49.58\scriptsize{$\pm$4.39} & 63.33 & 54.58\scriptsize{$\pm$5.12} & 66.67 & \textbf{45.14}\scriptsize{$\pm$0.70} \\
    ~ & \methodabbr{} (BS=16) & 2-2 & \textbf{67.29}\scriptsize{$\pm$3.38} & 76.67 & 49.79\scriptsize{$\pm$6.29} & \textbf{70.00} & \textbf{56.67}\scriptsize{$\pm$3.91} & \textbf{73.33} & 44.57\scriptsize{$\pm$0.40} \\
    
    % \midrule
    
    % ~ & - - & 16-16 & 69.17\scriptsize{$\pm$4.64} & 83.33 & 52.92\scriptsize{$\pm$6.65} & 66.67 \\
    % DeepSeek-LLaMA-70B & KIVI (BS=12,16) & 2-2 & \\
    % \cmidrule(lr){2-10}
    % (2$\times$A100-80G) & \methodabbr{} (BS=12) & 2/4-2/4 & 64.79\scriptsize{$\pm$5.77} & 86.67 \\
    % ~ & \methodabbr{} (BS=16) & 2-2 & 64.79\scriptsize{$\pm$4.85} & 83.33 & 47.08\scriptsize{$\pm$5.12} & 63.33 \\
    \bottomrule
  \end{tabular}
  }
  \vspace{-15pt}
\end{table}

\vspace{-3pt}
\subsection{Main Results}
\label{sec:exp-main}
\vspace{-5pt}

As illustrated in \cref{tab:main}, for long-CoT LLMs smaller than 10B, we compare \methodabbr{} with RotateKV, MiKV, and KIVI. 
% \methodabbr{} consistently outperforms the baseline methods across various benchmarks and memory budgets. 
For the 2-bit DeepSeek-R1-Distill-Qwen-7B, applying RotateKV or MiKV causes the model unable to generate meaningful responses.
% The average pass@1 across benchmarks drops to nearly 0\%.
The SOTA method KIVI also suffers from significant performance loss by up to 9\%.
\methodabbr{} outperforms KIVI by up to 8\% when applying uniform Fbit for each transformer block (batch size = 40). 
When the batch size is reduced to 32, each sample receives a larger memory budget.
However, this budget is still insufficient to apply uniform 4-bit quantization across all blocks.
As a result, KIVI is constrained to 2-bit quantization, underutilizing the available memory.
In contrast, \methodabbr{} leverages block-wise memory allocation to better utilize the larger memory, achieving an additional performance gain of up to 0.84\%.
For the 4-bit DeepSeek-R1-Distill-LLaMA-8B, %the MiKV and RotateKV can effectively preserve the performance under 4-bit quantization.
\methodabbr{} surpasses the SOTA methods by up to 6.5\% on AIME-2024, and even achieve better performance than the original LLM on mathematical benchmarks.
Besides, for LLMs smaller than 10B, the average voting accuracy of \methodabbr{} exceeds KIVI by up to 15.56\%, demonstrating greater stability of the proposed method.
We also compare \methodabbr{} with KIVI of different bit-widths in \cref{add:kivi}.

For larger long-CoT LLMs from 10B to 32B, we only compare the proposed \methodabbr{} with KIVI because MiKV and RotateKV fail to generate meaningful information under 2-bit quantization, as discovered in the 2-bit DeepSeek-R1-Distill-Qwen-7B.
As shown in \cref{tab:main}, \methodabbr{} also demonstrates superior performance compared to KIVI, improving average pass@1 and voting accuracy by up to 15.00\% and 17.78\% on various LLMs. 
Especially, for the DeepSeek-R1-Distill-Qwen-14B, KIVI causes a performance degradation of 21.87\% on CMIMC-2024, whereas \methodabbr{} has a significantly lower degradation of only 1.87\% and 2.91\% under batch sizes of 16 and 12, respectively.

For the 70B-level long-CoT LLM, we evaluate the 2-bit DeepSeek-R1-Distill-LLaMA-70B model on the AIME-2024 benchmark. The original 16-bit model achieves a pass@1 of 69.14\%.
When the KV Cache is quantized to 2-bit using KIVI, the pass@1 drops significantly to 51.88\%.
In contrast, the proposed \methodabbr{} enables the 2-bit model to achieve a much higher pass@1 of 64.79\% under both batch sizes of 12 and 16, outperforming the KIVI baseline by 12.91\%.

\vspace{-2pt}
\subsection{Efficiency Analysis}
\label{sec:exp-eff}
\vspace{-4pt}

% \begin{wraptable}{r}{0.5\linewidth}
%     \centering
%     \vspace{-8pt}
%     \caption{Time required before model inference. Alloc. and Calib are short for block-wise memory allocation and calibration with positional interpolation.}
%     \begin{tabular}{cccc}
%         \toprule
%         Model & Alloc. & Calib. \\
%         \midrule
%         DeepSeek-LLaMA-8B & 4 min & 17 min \\
%         DeepSeek-Qwen-32B & 13 min & 44 min \\
%         DeepSeek-LLaMA-70B & 26 min & 93 min \\
%         \bottomrule
%     \end{tabular}
%     \label{tab:pre-infer-time}
% \end{wraptable}

% \textbf{Pre-inference efficiency.} 
% Before the inference process, \methodabbr{} performs block-wise memory allocation and calibration with positional interpolation.
% The time required for these two processes is reported in \cref{tab:pre-infer-time}.
% Note that both steps are conducted offline, which means that they only need to be executed once and the results can be reused across multiple deployments.

% \textbf{Inference efficiency.} 

\begin{wraptable}{r}{0.55\linewidth}
    \centering
    \vspace{-10pt}
    \caption{The throughput (in tokens/s) across different quantization methods and output lengths.}
    % \vspace{-5pt}
    \resizebox{\linewidth}{!}{
    \begin{tabular}{ccccc}
        \toprule
        \multirow{2}{*}{Model} & Quantization & \multicolumn{3}{c}{Output Length}\\
        \cmidrule(lr){3-5}
        % ~ & Method & 16384 & 24576 & 32768 \\
        ~ & Method & 16K & 24K & 32K \\
        \midrule
      
        \multirow{3}{*}{\makecell[l]{DeepSeek-\\Qwen-7B}} & -- & 101.40 & 65.69 & 52.06 \\
        ~ & KIVI & 506.72 & 352.44 & 284.10 \\
        ~ & \methodabbr{} & 424.72 & 323.48 & 269.51 \\

        \midrule

        \multirow{3}{*}{\makecell[l]{DeepSeek-\\Qwen-32B}} & -- & 12.34 & 10.35 & 8.92 \\
        ~ & KIVI & 35.65 & 33.28 & 31.59 \\
        ~ & \methodabbr{} & 33.74 & 32.15 & 30.81 \\
        
        \bottomrule
    \end{tabular}
    }
    \label{tab:efficiency}
    \vspace{-5pt}
\end{wraptable}
We evaluate 7B and 32B long-CoT LLMs on an A100-80G GPU, comparing the throughput of \methodabbr{} (Fbit=2) against the original 16-bit LLMs and the 2-bit KIVI baseline.
% For each method, the batch size is set to the maximum value that allows the GPU to accommodate KV Cache for 32K tokens.
We adopt the official settings of KIVI~\cite{kivi}, using its inference engine and 4/2-bit CUDA kernels for efficiency evaluation.
Besides, we implement 16/8-bit CUDA kernels and bit-width shrinking kernels to support \methodabbr{}.
To fully utilize the A100-80G memory, we set the batch sizes of the original 7B and 32B models to 18 and 1, respectively, while the quantized models allow larger batch sizes of 110 and 4.
%% The quantization bit-width of KIVI and Fbit of \methodabbr{} are both set to 2-bit.
% \todo{Which inference engine was used? Whose CUDA kernels were adopted? What did we implement ourselves?}

As shown in \cref{tab:efficiency}, across different model sizes and output lengths, \methodabbr{} achieves a 2.73–5.18$\times$ throughput improvement over the original 16-bit LLMs.
% Although the throughput of \methodabbr{} is XX–XX\% lower than that of KIVI, it delivers higher accuracy.
% Compared with KIVI, the throughput of \methodabbr{} remains in the same order of magnitude, with a slight reduction primarily caused by the use of higher bit-widths during inference.
Compared with KIVI, the throughput of \methodabbr{} is at a similar level, with a slight reduction primarily due to the use of higher bit-widths during inference.
Notably, the overhead of bit-width shrinking is negligible, as it is triggered only when memory is fully utilized.
\textbf{Overall, \methodabbr{} incurs a throughput degradation of 2.45–16.18\% compared to KIVI but achieves a substantial relative accuracy improvement of 10.57–23.48\%.}
% We also test the time cost of block-wise memory allocation and calibration with positional interpolation.
% See \cref{sec:pre-inference-time} for details.
To further evaluate the efficiency of the quantization procedure, we measure the latency of block-wise memory allocation and calibration with positional interpolation. % to show the speed of the quantization procedure.
As shown in \cref{sec:pre-inference-time}, both the 7B and 32B LLMs complete these procedures within one hour using \methodabbr{}.
% As shown in \cref{sec:pre-inference-time}, for both the 7B and 32B models, \methodabbr{} completes these two procedures within one hour.

% \todo{block-wise memory allocation time. For different LLM, give us some speed value @CVXPY.}

\vspace{-5pt}
\subsection{Ablation Studies}
\label{sec:exp-ablation}
In this section, we conduct detailed ablation studies to show the effect of bit-wise shrinking strategies introduced in \cref{sec:method-token}, the sensitivity of different transformer blocks discussed in \cref{sec:method-layer}, and the effectiveness of the positional interpolation discussed in \cref{sec:method-calib}.

\subsubsection{The Effect of Bit-width Shrinking Strategies}
\label{sec:exp-ablation-pm}

\begin{table}[thbp]
    \caption{Ablation results of different bit-width shrinking strategies.}
    \vspace{-5pt}
    \centering
    \resizebox{0.8\linewidth}{!}{
    \begin{tabular}{ccccc}
    \toprule
        \multirow{2}{*}{Model} & \multirow{2}{*}{Bit-width Shrinking Strategy} & Bit-width & \multicolumn{2}{c}{AIME-2024}\\
        \cmidrule(lr){3-3} \cmidrule(lr){4-5}
        ~ & ~ & (K-V) & pass@1 & Voting \\
        \midrule 
        
         \multirow{4}{*}{DeepSeek-LLaMA-8B}& - - & 16-16 & 44.17 & 66.67 \\
         ~ & Direct Right Shift & 4-4 & 12.08 & 23.33 \\
         ~ & Modified Right Shift & 4-4 & 28.75 & 46.67 \\
         ~ & Equivalent Right Shift (Ours) & 4-4 & 38.33 & 66.67 \\
    \bottomrule
    \end{tabular}
    }
    % \vspace{-10pt}
    \label{tab:ablation-progressive}
\end{table}

We compare three different bit-width shrinking strategies for reducing the KV Cache from $2b$-bit to $b$-bit.
Specifically, $b$ can be 8, 4, or 2, corresponding to shrinking the KV Cache from 16-bit to 8-bit, 8-bit to 4-bit, and 4-bit to 2-bit, respectively.

\begin{figure}[t]
    \centering
    \includegraphics[width=1\linewidth]{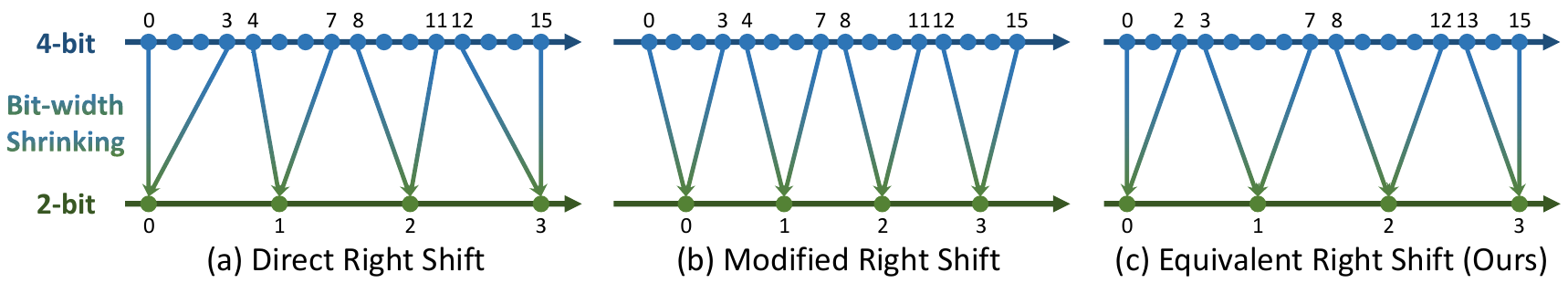}
    \vspace{-18pt}
    \caption{Different bit-width shrinking strategies when the bit-width is reduced from 4-bit to 2-bit.}
    \label{fig:shift}
    \vspace{-15pt}
\end{figure}

(1) \textbf{Direct Right Shift}: By directly right-shifting by $b$ bits, only the higher $b$ bits of the original $2b$-bit value are retained.
% this strategy simply right shifts the quantized KV Cache by $b$-bit, retaining only the most significant $b$-bit digits. 这半句话是啥意思，什么叫最显著的 $b$-bit ？
As shown in \cref{fig:shift} (a), to preserve the dynamic range of the quantized values, we keep the zero point unchanged ($Z_b = Z_{2b}$) and increase the scaling factor to $S_b = (2^b + 1) S_{2b}$ to compensate for the magnitude reduction caused by the right-shift operation.
% However, it leads to a discrepancy between direct $b$-bit quantization and progressive quantization. 这句话没有信息量，没有解释原因。删掉，不能出现这种武断的说法。

(2) \textbf{Modified Right Shift}: This strategy also uses $b$-bit right shifting strategy to perform the bit-width shrinking.
However, instead of keeping the dynamic range unchanged, this strategy aims to ensure that quantization levels sharing the same upper $b$ bits before the shift can have their mean values directly mapped to the lower bit-width representation, as demonstrated in \cref{fig:shift} (b).
To achieve this, we change the scaling factor by $S_b=2^b\cdot S_{2b}$ and zero point by $Z_b=Z_{2b}+\frac{1}{2}(S_b-S_{2b})$.

(3) \textbf{Equivalent Right Shift (in \cref{sec:method-token})}: As shown in \cref{fig:shift} (c), this strategy is equivalent to directly de-quantizing the $2b$-bit KV Cache and then quantizing it to $b$-bit.
% The detailed proof of the equivalence is shown in \cref{add:proof}.

We evaluate the above three bit-width shrinking strategies on the AIME-2024 benchmark with DeepSeek-R1-Distill-LLaMA-8B.
As shown in \cref{tab:ablation-progressive}, both the Direct Right Shift and Modified Right Shift strategies result in significant performance degradation, reducing the pass@1 by 32.09\% and 15.42\%, respectively.
In contrast, the Equivalent Right Shift demonstrates a notable improvement over the other two strategies, increasing the pass@1 by 26.25\% and 9.58\%, and maintaining a lossless voting accuracy.
Therefore, we adopt the Equivalent Right Shift strategy in \methodabbr{}.
\subsubsection{The Sensitivity of Different Transformer Blocks}
\label{sec:exp-ablation-sen}

\begin{figure}[htbp]
    \begin{subfigure}[b]{1\linewidth}
        \centering
        \includegraphics[width=\linewidth]{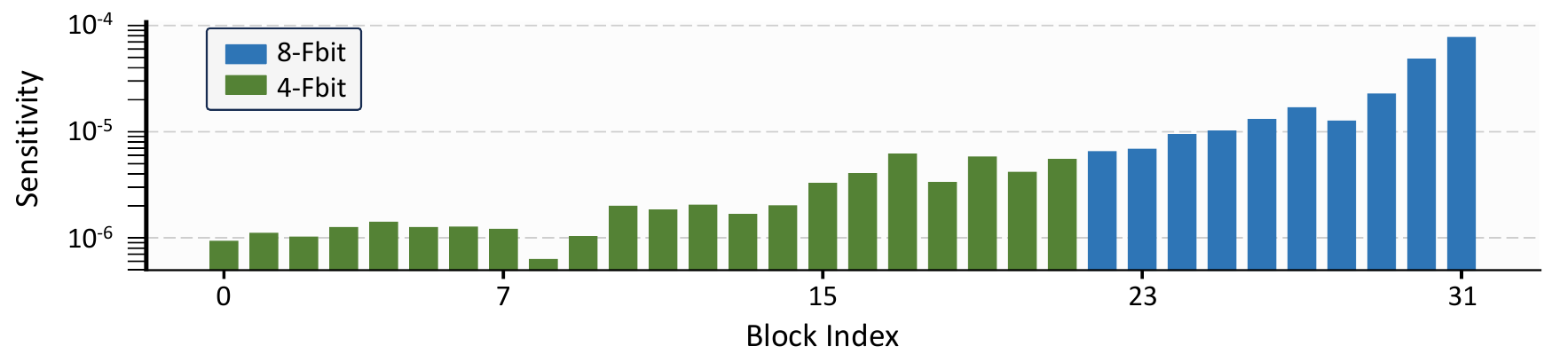}
        \caption{DeepSeek-R1-Distill-LLaMA-8B}
        \label{fig:subfig-llama-8b}
    \end{subfigure}
    
    \centering
    \begin{subfigure}[b]{1\linewidth}
        \centering
        \includegraphics[width=\linewidth]{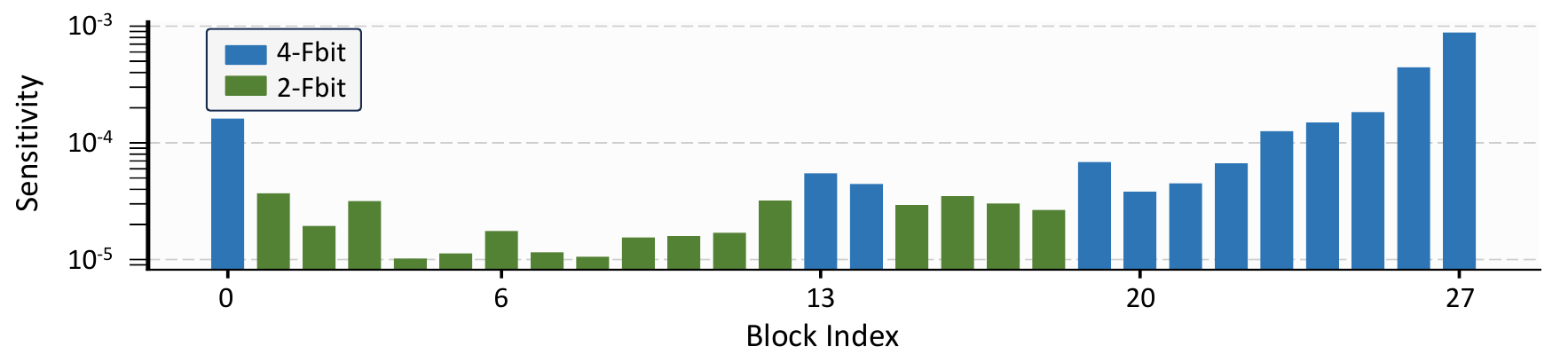}
        \caption{DeepSeek-R1-Distill-Qwen-7B}
        \label{fig:subfig-qwen-7b}
    \end{subfigure}

    \caption{Sensitivity to quantization of KV Cache in different transformer blocks. Different colors represents different memory budgets.}
    \label{fig:sensitivity-small}
    \vspace{-3pt}
\end{figure}

We analyze the sensitivity and the memory allocation results across different models. 
For models with parameter size less than 10B, as shown in \cref{fig:sensitivity-small}, we observe that the deeper blocks tend to be more sensitive to quantization and receive a larger memory budget for the KV Cache.
In addition, in the DeepSeek-R1-Distill-Qwen-7B model, the first block is more sensitive than the other shallow blocks. 
% A detailed analysis of the data distribution indicates that, in this model, the Key Cache in the first block exhibits markedly higher outlier magnitudes than other blocks, leading to a significantly increased quantization error.
Our memory allocation strategy accurately captures this feature, assigning a higher memory budget to the first block accordingly.
The block sensitivity of larger models is detailed in \cref{add:exp-sen}.

\subsubsection{The Effect of Positional Interpolation}
\label{sec:exp-ablation-layer}
We evaluate the long-CoT performance across varying lengths of calibration data and position scaling factor $s$.
We utilize the DeepSeek-R1-Distill-LLaMA-8B to generate four responses for each problem in the AIME-2024-I dataset.
As shown in \cref{tab:ablation-pi}, when the calibration sequence length is set to 2,048, applying positional interpolation with $s = 4$ improves pass@1 by 1.66\% compared to not using positional interpolation, achieving accuracy comparable to that obtained using calibration sequences of 8,192 tokens.
We also observe that when $s$ increases to 16, positional interpolation may lead to performance degradation.
This indicates that the computational savings of positional interpolation are not unlimited, and overly aggressive scaling can indeed performance drop.

\begin{table}[h]
    \centering
    \caption{Ablation results of different calibration sequence lengths and position scaling factors.}
    % \vspace{-5pt}
    \resizebox{0.8\linewidth}{!}{
    \begin{tabular}{cccccc}
        \toprule
        \multirow{2}{*}{Model} & \multirow{2}{*}{\makecell{Calibration Sequence\\Length}} & \multirow{2}{*}{\makecell{Position Scaling\\Factor}} & \multirow{2}{*}{\makecell{Effective\\Length}} & \multicolumn{2}{c}{AIME-2024-I} \\
        \cmidrule(lr){5-6}
        ~ & ~ & ~ & ~ & pass@1 & Voting \\
        \midrule
        \multirow{4}{*}{\makecell[l]{DeepSeek-\\LLaMA-8B}} & 2,048 & 1 & 2,048 & 46.67 & 60.00 \\
        ~ & 2,048 & 4 & 8,192 & 48.33 & 60.00 \\
        ~ & 2,048 & 16 & 32,768 & 46.67 & 53.33 \\
        ~ & 8,192 & 1 & 8,192 & 48.33 & 60.00 \\
        \bottomrule
    \end{tabular}
    }
    \vspace{-15pt}
    \label{tab:ablation-pi}
\end{table}

\section{Conclusion}
\label{sec:conclusion}

In this paper, we introduce Progressive Mixed-precision KV Cache Quantization (\methodabbr{}), a post-training KV Cache quantization method designed for long-CoT LLMs.
To reduce the large cumulative error caused by uniform bit-width quantization, we design progressive quantization and block-wise memory allocation techniques.
To increase the effective calibration length without incurring additional overhead, we propose a new calibration strategy with positional interpolation.
Extensive experiments and ablation studies demonstrate the effectiveness of the proposed \methodabbr{} and each proposed technique.
Overall, the proposed \methodabbr{} significantly outperforms SOTA baselines by up to 8\% on reasoning-related mathematics and coding benchmarks and achieves 2.73–5.18× throughput compared to the original 16-bit LLMs.

\section*{Acknowledgement}
This work was supported by National Natural Science Foundation of China (No. 62325405, 62104128, U19B2019, U21B2031, 61832007, 62204164, 92364201), Tsinghua EE Xilinx AI Research Fund, and Beijing National Research Center for Information Science and Technology (BNRist).
We thank for all the support from Infinigence-AI.

\section*{Ethics Statement}
This work focuses on reducing the substantial overhead caused by the linearly growing KV cache in long-context processing through KV Cache quantization.
On the one hand, the proposed \methodabbr{} better preserves model accuracy after low-precision KV cache quantization, making it more accessible for cost-constrained institutions, individuals, and application scenarios.
On the other hand, as a lossy compression technique, quantization can introduce distribution shifts and performance degradation, potentially leading to increased hallucinations or instruction-following failures.
Therefore, additional caution and oversight are required during deployment.

\section*{Reprodicibility Statement}
% \todo{add Reprodicibility Statement}
We describe the calibration and evaluation datasets, as well as the data processing procedures, in \cref{sec:exp-datasets}. 
All datasets and models used in our experiments are publicly available. 
Detailed information on the quantization bit-widths and batch sizes used for each long-CoT LLM is also provided in \cref{sec:exp-setups-budget}. 
To facilitate reproducibility, we also release our source code along with detailed guidelines.

\bibliography{iclr2026_conference}

@article{deepseek-r1,
  title={Deepseek-r1: Incentivizing reasoning capability in llms via reinforcement learning},
  author={Guo, Daya and Yang, Dejian and Zhang, Haowei and Song, Junxiao and Zhang, Ruoyu and Xu, Runxin and Zhu, Qihao and Ma, Shirong and Wang, Peiyi and Bi, Xiao and others},
  journal={arXiv preprint arXiv:2501.12948},
  year={2025}
}

@misc{qwq32b,
    title = {QwQ-32B: Embracing the Power of Reinforcement Learning},
    url = {https://qwenlm.github.io/blog/qwq-32b/},
    author = {Qwen Team},
    month = {March},
    year = {2025}
}

@misc{openai-o1,
    title = {Introducing OpenAI o1},
    url = {https://openai.com/o1/},
    author = {OpenAI},
    month = {September},
    year = {2024}
}

@misc{aime,
    title = {American Invitational Mathematics Examination},
    url = {https://artofproblemsolving.com/},
    author = {AIME},
    year = {2025}
}

@misc{cmimc,
    title = {Carnegie Mellon Informatics and Mathematics Competition},
    url = {https://cmimc.math.cmu.edu/math},
    author = {CMIMC},
    year = {2025}
}

@article{mqa,
  title={Fast transformer decoding: One write-head is all you need},
  author={Shazeer, Noam},
  journal={arXiv preprint arXiv:1911.02150},
  year={2019}
}

@article{gqa,
  title={Gqa: Training generalized multi-query transformer models from multi-head checkpoints},
  author={Ainslie, Joshua and Lee-Thorp, James and De Jong, Michiel and Zemlyanskiy, Yury and Lebr{\'o}n, Federico and Sanghai, Sumit},
  journal={arXiv preprint arXiv:2305.13245},
  year={2023}
}

@article{mla,
  title={Deepseek-v2: A strong, economical, and efficient mixture-of-experts language model},
  author={Liu, Aixin and Feng, Bei and Wang, Bin and Wang, Bingxuan and Liu, Bo and Zhao, Chenggang and Dengr, Chengqi and Ruan, Chong and Dai, Damai and Guo, Daya and others},
  journal={arXiv preprint arXiv:2405.04434},
  year={2024}
}

@article{cvxpy,
  author  = {Steven Diamond and Stephen Boyd},
  title   = {{CVXPY}: {A} {P}ython-embedded modeling language for convex optimization},
  journal = {Journal of Machine Learning Research},
  year    = {2016},
  volume  = {17},
  number  = {83},
  pages   = {1--5},
}

@article{awq,
  title={Awq: Activation-aware weight quantization for on-device llm compression and acceleration},
  author={Lin, Ji and Tang, Jiaming and Tang, Haotian and Yang, Shang and Chen, Wei-Ming and Wang, Wei-Chen and Xiao, Guangxuan and Dang, Xingyu and Gan, Chuang and Han, Song},
  journal={Proceedings of Machine Learning and Systems},
  volume={6},
  pages={87--100},
  year={2024}
}

@article{kivi,
  title={Kivi: A tuning-free asymmetric 2bit quantization for kv cache},
  author={Liu, Zirui and Yuan, Jiayi and Jin, Hongye and Zhong, Shaochen and Xu, Zhaozhuo and Braverman, Vladimir and Chen, Beidi and Hu, Xia},
  journal={arXiv preprint arXiv:2402.02750},
  year={2024}
}

@article{rotatekv,
  title={RotateKV: Accurate and Robust 2-Bit KV Cache Quantization for LLMs via Outlier-Aware Adaptive Rotations},
  author={Su, Zunhai and Chen, Zhe and Shen, Wang and Wei, Hanyu and Li, Linge and Yu, Huangqi and Yuan, Kehong},
  journal={arXiv preprint arXiv:2501.16383},
  year={2025}
}

@article{skvq,
  title={Skvq: Sliding-window key and value cache quantization for large language models},
  author={Duanmu, Haojie and Yuan, Zhihang and Li, Xiuhong and Duan, Jiangfei and Zhang, Xingcheng and Lin, Dahua},
  journal={arXiv preprint arXiv:2405.06219},
  year={2024}
}

@article{mklv,
  title={More for Keys, Less for Values: Adaptive KV Cache Quantization},
  author={Hariri, Mohsen and Nguyen, Lam and Chen, Sixu and Zhong, Shaochen and Wang, Qifan and Hu, Xia and Han, Xiaotian and Chaudhary, Vipin},
  journal={arXiv preprint arXiv:2502.15075},
  year={2025}
}

@article{wkvquant,
  title={Wkvquant: Quantizing weight and key/value cache for large language models gains more},
  author={Yue, Yuxuan and Yuan, Zhihang and Duanmu, Haojie and Zhou, Sifan and Wu, Jianlong and Nie, Liqiang},
  journal={arXiv preprint arXiv:2402.12065},
  year={2024}
}

@article{mikv,
  title={No token left behind: Reliable kv cache compression via importance-aware mixed precision quantization},
  author={Yang, June Yong and Kim, Byeongwook and Bae, Jeongin and Kwon, Beomseok and Park, Gunho and Yang, Eunho and Kwon, Se Jung and Lee, Dongsoo},
  journal={arXiv preprint arXiv:2402.18096},
  year={2024}
}

@article{streamingllm,
  title={Efficient streaming language models with attention sinks},
  author={Xiao, Guangxuan and Tian, Yuandong and Chen, Beidi and Han, Song and Lewis, Mike},
  journal={arXiv preprint arXiv:2309.17453},
  year={2023}
}

@article{h2o,
  title={H2o: Heavy-hitter oracle for efficient generative inference of large language models},
  author={Zhang, Zhenyu and Sheng, Ying and Zhou, Tianyi and Chen, Tianlong and Zheng, Lianmin and Cai, Ruisi and Song, Zhao and Tian, Yuandong and R{\'e}, Christopher and Barrett, Clark and others},
  journal={Advances in Neural Information Processing Systems},
  volume={36},
  pages={34661--34710},
  year={2023}
}

@article{moa,
  title={Moa: Mixture of sparse attention for automatic large language model compression},
  author={Fu, Tianyu and Huang, Haofeng and Ning, Xuefei and Zhang, Genghan and Chen, Boju and Wu, Tianqi and Wang, Hongyi and Huang, Zixiao and Li, Shiyao and Yan, Shengen and others},
  journal={arXiv preprint arXiv:2406.14909},
  year={2024}
}

@article{qserve,
  title={QServe: W4A8KV4 Quantization and System Co-design for Efficient LLM Serving},
  author={Lin*, Yujun and Tang*, Haotian and Yang*, Shang and Zhang, Zhekai and Xiao, Guangxuan and Gan, Chuang and Han, Song},
  journal={arXiv preprint arXiv:2405.04532},
  year={2024}
}

@inproceedings{llm-mq,
  title={Llm-mq: Mixed-precision quantization for efficient llm deployment},
  author={Li, Shiyao and Ning, Xuefei and Hong, Ke and Liu, Tengxuan and Wang, Luning and Li, Xiuhong and Zhong, Kai and Dai, Guohao and Yang, Huazhong and Wang, Yu},
  booktitle={NeurIPS 2023 Efficient Natural Language and Speech Processing Workshop},
  pages={1--5},
  year={2023}
}

@inproceedings{mixdq,
  title={Mixdq: Memory-efficient few-step text-to-image diffusion models with metric-decoupled mixed precision quantization},
  author={Zhao, Tianchen and Ning, Xuefei and Fang, Tongcheng and Liu, Enshu and Huang, Guyue and Lin, Zinan and Yan, Shengen and Dai, Guohao and Wang, Yu},
  booktitle={European Conference on Computer Vision},
  pages={285--302},
  year={2024},
  organization={Springer}
}

@inproceedings{intactkv,
  title={IntactKV: Improving Large Language Model Quantization by Keeping Pivot Tokens Intact},
  author={Liu, Ruikang and Bai, Haoli and Haokun, LIN and Li, Yuening and Gao, Han and Xu, Zhengzhuo and Hou, Lu and Yao, Jun and Yuan, Chun},
  booktitle={The 62nd Annual Meeting of the Association for Computational Linguistics (ACL 2024)},
  year={2024}
}

@article{pi,
  title={Extending context window of large language models via positional interpolation},
  author={Chen, Shouyuan and Wong, Sherman and Chen, Liangjian and Tian, Yuandong},
  journal={arXiv preprint arXiv:2306.15595},
  year={2023}
}

@article{redpajama,
	title   = {RedPajama: an Open Dataset for Training Large Language Models},
	author  = {Maurice Weber and Daniel Y. Fu and Quentin Anthony and Yonatan Oren and Shane Adams and Anton Alexandrov and Xiaozhong Lyu and Huu Nguyen and Xiaozhe Yao and Virginia Adams and Ben Athiwaratkun and Rahul Chalamala and Kezhen Chen and Max Ryabinin and Tri Dao and Percy Liang and Christopher Ré and Irina Rish and Ce Zhang},
	journal = {NeurIPS Datasets and Benchmarks Track},
	year    = 2024,
}

@article{livecodebench,
  title={Livecodebench: Holistic and contamination free evaluation of large language models for code},
  author={Jain, Naman and Han, King and Gu, Alex and Li, Wen-Ding and Yan, Fanjia and Zhang, Tianjun and Wang, Sida and Solar-Lezama, Armando and Sen, Koushik and Stoica, Ion},
  journal={arXiv preprint arXiv:2403.07974},
  year={2024}
}

@article{rope,
  title={Roformer: Enhanced transformer with rotary position embedding},
  author={Su, Jianlin and Ahmed, Murtadha and Lu, Yu and Pan, Shengfeng and Bo, Wen and Liu, Yunfeng},
  journal={Neurocomputing},
  volume={568},
  pages={127063},
  year={2024},
  publisher={Elsevier}
}

@misc{opencompass,
    title={OpenCompass: A Universal Evaluation Platform for Foundation Models},
    author={OpenCompass Contributors},
    howpublished = {\url{https://github.com/open-compass/opencompass}},
    year={2023}
}

@article{ifeval,
  title={Instruction-following evaluation for large language models},
  author={Zhou, Jeffrey and Lu, Tianjian and Mishra, Swaroop and Brahma, Siddhartha and Basu, Sujoy and Luan, Yi and Zhou, Denny and Hou, Le},
  journal={arXiv preprint arXiv:2311.07911},
  year={2023}
}

@article{kvtuner,
  title={Kvtuner: Sensitivity-aware layer-wise mixed-precision kv cache quantization for efficient and nearly lossless llm inference},
  author={Li, Xing and Xing, Zeyu and Li, Yiming and Qu, Linping and Zhen, Hui-Ling and Liu, Wulong and Yao, Yiwu and Pan, Sinno Jialin and Yuan, Mingxuan},
  journal={arXiv preprint arXiv:2502.04420},
  year={2025}
}

@article{quest,
  title={Quest: Query-aware sparsity for efficient long-context llm inference},
  author={Tang, Jiaming and Zhao, Yilong and Zhu, Kan and Xiao, Guangxuan and Kasikci, Baris and Han, Song},
  journal={arXiv preprint arXiv:2406.10774},
  year={2024}
}
\bibliographystyle{iclr2026_conference}

\newpage
\appendix
\section{The Use of Large Language Models (LLMs)}
In this paper, LLMs are only used to assist in polishing the writing of this paper.
The technical content, experiments, and conclusions are entirely conceived and conducted by the authors.

\section{Additional Details of Evaluation}
\label{add:exp-eval}

\subsection{Introduction of Datasets}
\label{add:dataset}
\textbf{American Invitational Mathematics Examination (AIME)}~\cite{aime} is a mathematics competition for high school students. It contains 30 challenging problems each year, designed to assess mathematical problem-solving skills across various topics, including algebra, combinatorics, geometry, number theory, and other subjects covered in high school curricula.

\textbf{Carnegie Mellon Informatics and Mathematics Competition (CMIMC)}~\cite{cmimc} is an annual mathematics contest for high school students, hosted by students from Carnegie Mellon University.
The competition contains problems of algebra, combinatorics, and geometry, with each category including ten standard problems along with one tiebreaker.
Our model evaluation focuses on the standard problem sets.

\textbf{LiveCodeBench}~\cite{livecodebench} is an extensive and continuously updated benchmark designed to evaluate the performance of LLMs in coding tasks.
It continually gathers new problems from competition platforms.
The benchmark encompasses four distinct scenarios: code generation, automated code repair, code execution, and prediction of test outputs.
In our experiments, we focus specifically on the code generation scenario.

\textbf{IFEval}~\cite{ifeval} is a benchmark proposed to systematically evaluate the ability of LLMs to follow natural language instructions. 
The dataset contains 541 prompts, each annotated with one or more verifiable instruction types such as word-count constraints, keyword frequency, formatting requirements, or prohibitions on certain symbols. 
These instructions were deliberately designed to be automatically checkable, enabling objective and reproducible evaluation without the need for human annotators.

\section{Additional Experiments}
\label{add:exp}

\subsection{The Sensitivity of different Transformer Blocks}
\label{add:exp-sen}

% \begin{figure}[htbp]
%     \begin{subfigure}[b]{1\linewidth}
%         \centering
%         \includegraphics[width=\linewidth]{figures/sensitivity-llama-8b.pdf}
%         \caption{DeepSeek-R1-Distill-LLaMA-8B}
%         \label{fig:subfig-llama-8b}
%     \end{subfigure}
    
%     \centering
%     \begin{subfigure}[b]{1\linewidth}
%         \centering
%         \includegraphics[width=\linewidth]{figures/sensitivity-qwen-7b.pdf}
%         \caption{DeepSeek-R1-Distill-Qwen-7B}
%         \label{fig:subfig-qwen-7b}
%     \end{subfigure}

%     \caption{Sensitivity to quantization of KV Cache in different transformer blocks. Different colors represents different memory budgets.}
%     \label{fig:sensitivity}
% \end{figure}

\begin{figure}[htbp]
    \begin{subfigure}[b]{1\linewidth}
        \centering
        \includegraphics[width=\linewidth]{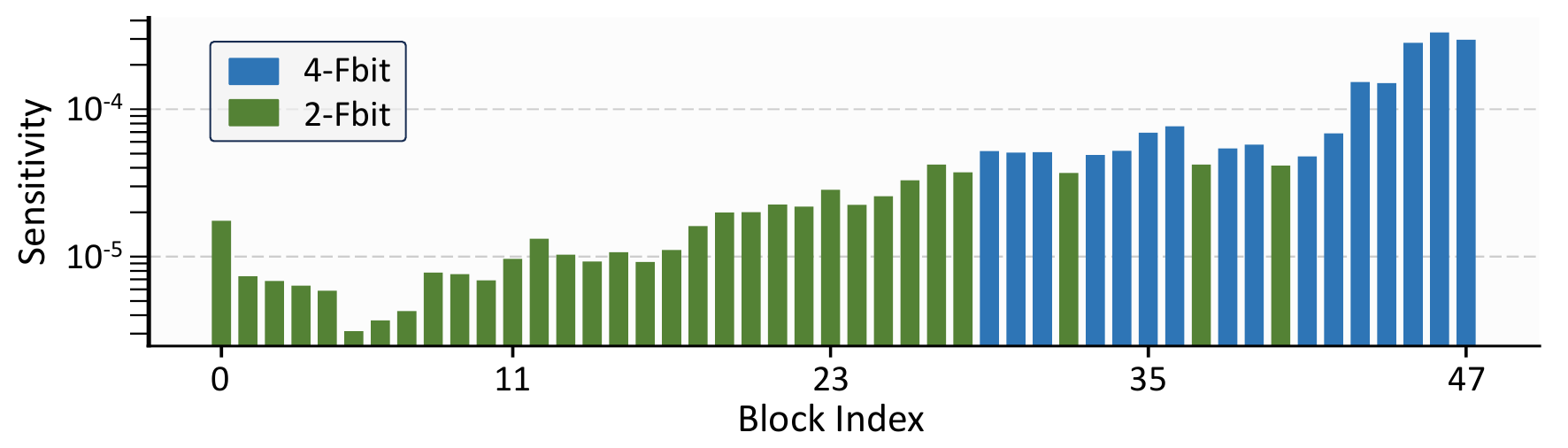}
        \caption{DeepSeek-R1-Distill-Qwen-14B}
        \label{fig:subfig-qwen-14b}
    \end{subfigure}
    
    \centering
    \begin{subfigure}[b]{1\linewidth}
        \centering
        \includegraphics[width=\linewidth]{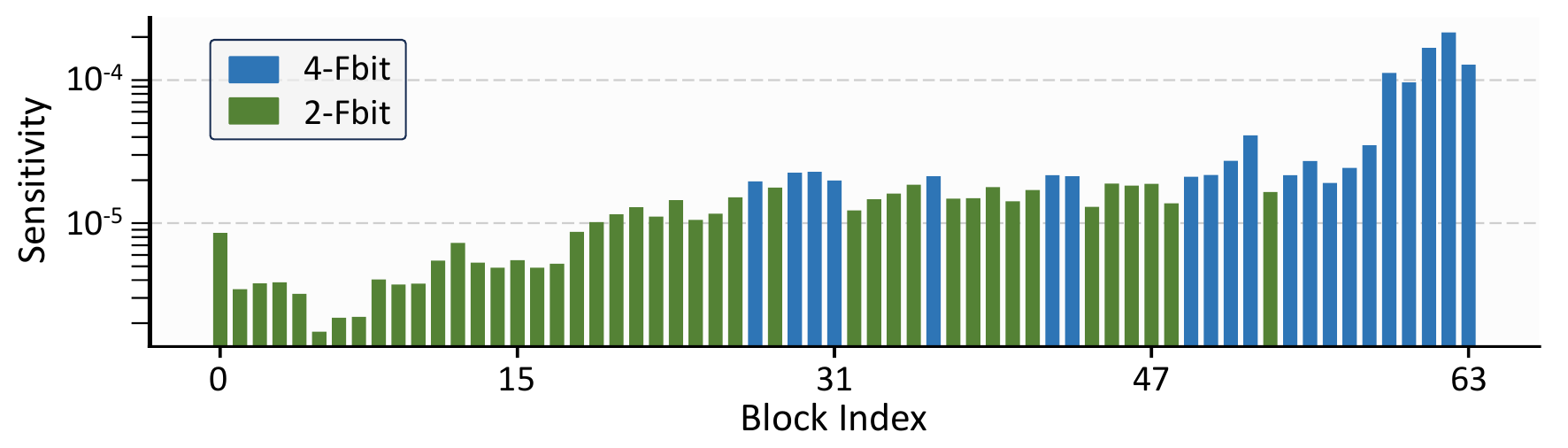}
        \caption{DeepSeek-R1-Distill-Qwen-32B}
        \label{fig:subfig-qwen-32b}
    \end{subfigure}

    \centering
    \begin{subfigure}[b]{1\linewidth}
        \centering
        \includegraphics[width=\linewidth]{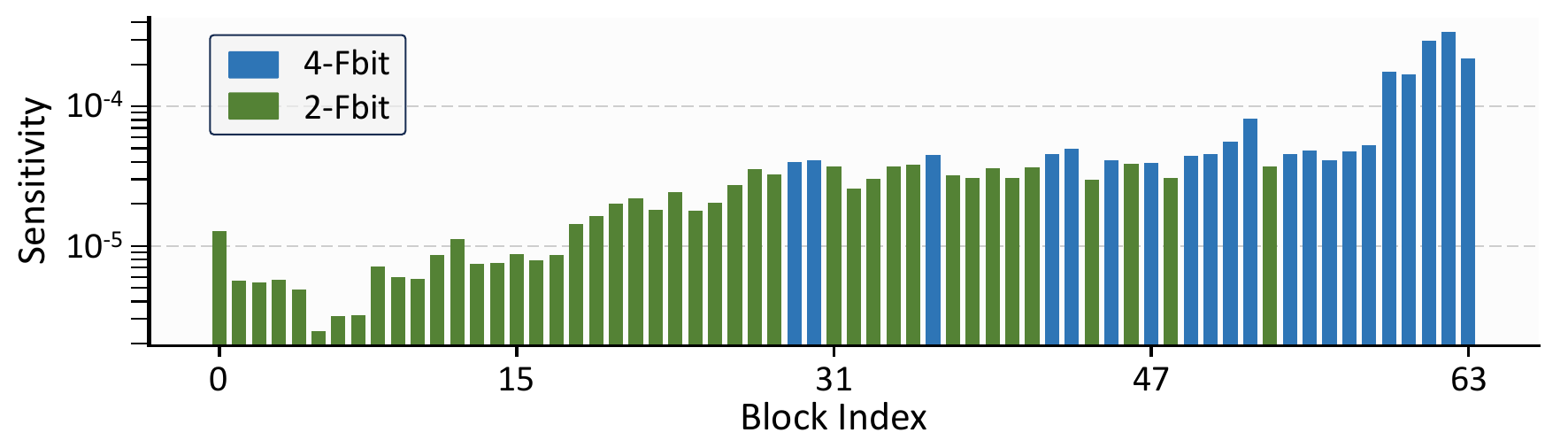}
        \caption{DeepSeek-R1-Distill-QwQ-32B}
        \label{fig:subfig-qwq-32b}
    \end{subfigure}

    \centering
    \begin{subfigure}[b]{1\linewidth}
        \centering
        \includegraphics[width=\linewidth]{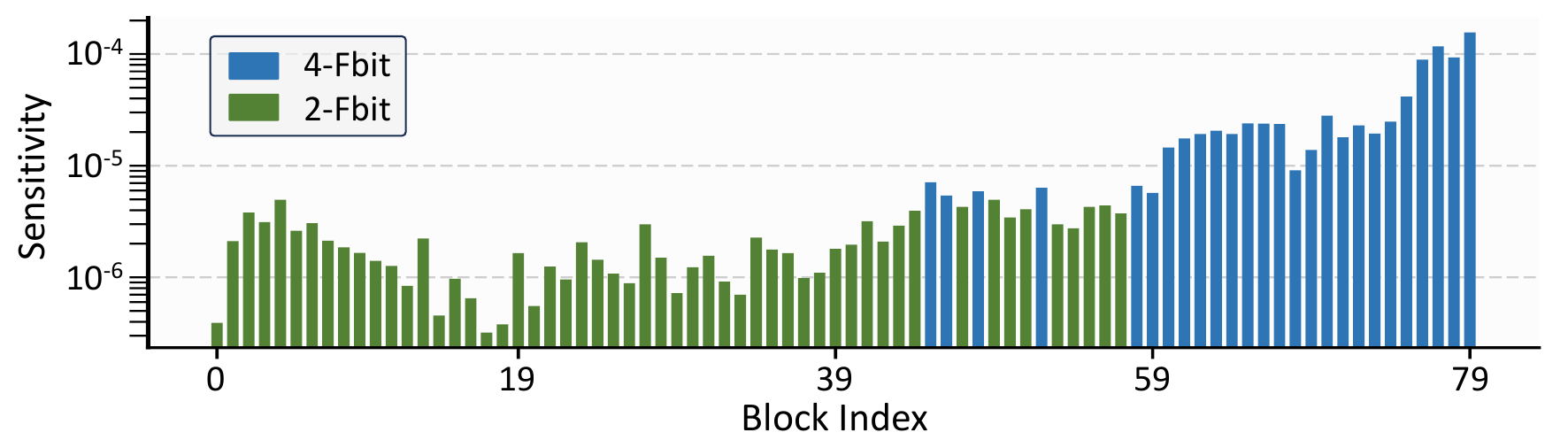}
        \caption{DeepSeek-R1-Distill-LLaMA-70B}
        \label{fig:subfig-llama-70b}
    \end{subfigure}

    \caption{Sensitivity to quantization of KV Cache in different transformer blocks. Different colors represents different memory budgets.}
    \label{fig:app-sensitivity}
\end{figure}

% For each model, we select two batch sizes.
% The larger batch size corresponds to a tight memory budget that exactly accommodates the KV Cache of the lower optional bit-width, except for the DeepSeek-R1-Distill-Qwen-7B model.
% The smaller batch size corresponds to a larger memory budget, allowing for the possibility of sensitivity-based memory allocation as described in \cref{sec:method-layer}. 
% % If we use a uniform memory allocation under the smaller batch size, the model's performance generally matches that of the larger batch size, except in the case of the DeepSeek-R1-Distill-Qwen-7B model. 
% As presented in \cref{tab:main}, for the DeepSeek-R1-Distill-LLaMA-8B model, employing the sensitivity-based memory allocation strategy improves the average pass@1 and voting accuracy by 0.92\% and 2.22\%, respectively, compared to uniform allocations.
% On large long-CoT LLMs, as shown in \cref{tab:main},  our memory allocation strategy will also improve the average pass@1 by up to 2.57\%.

For models with parameter size over 10B, as shown in \cref{fig:app-sensitivity}, KV Cache in deeper blocks tend to be more sensitive than shallower blocks.
We also observe that for the Qwen-based models, the first block exhibits a large sensitivity.
In particular, the sensitivity of the first block is the largest among the first fifteen blocks in different Qwen-based models.
This phenomenon is not observed in the LLaMA-based models.

\subsection{Performance in Short-generation-context Tasks}
\label{add:non-reasoning}
To verify the scalability of \methodabbr{} to short-generation-context tasks, we evaluate it on IFEval~\cite{ifeval}, an instruction-following benchmark.
We follow the experimental setup described in \cref{sec:exp-setups} and adopt the evaluation metrics provided by OpenCompass~\cite{opencompass}.
Compared to reasoning benchmarks in \cref{tab:main}, non-reasoning tasks are less challenging and generally involve much shorter outputs.
For instance, the average output length of the DeepSeek-Qwen-7B model is 13,904 tokens on AIME-2024 but only 1,182 tokens on IFEval.
As shown in \cref{tab:non-reasoning}, although our method is not specifically designed for short-output scenarios, it outperforms KIVI and achieves accuracy comparable to the original 16-bit models.

\begin{table}
  \caption{Results of long-CoT Language Models on non-reasoning benchmarks with SOTA KV Cache quantization methods. ``BS'' is short for ``batch size''.}
  \label{tab:non-reasoning}
  \centering
  \resizebox{\linewidth}{!}{
  \begin{tabular}{llccccc}
    \toprule
    Models & Quantization & Bit-width & \multicolumn{4}{c}{IFEval} \\
    \cmidrule(lr){3-3}\cmidrule(lr){4-7}
    (Target GPU) & Methods & (K-V) & Prompt Strict & Prompt Loose & Instruct Strict & Instruct Loose \\

    \midrule
    
    ~ & - - & 16-16 & 58.77 & 68.50 & 63.27 & 72.60 \\
    ~ & RotateKV (BS=32,40) & 2-2 & 0.00 & 0.00 & 0.00 & 0.00 \\ 
    DeepSeek- & MiKV (BS=32) & 2/16-2/16 & 57.30 & 66.17 & 61.18 & 70.06 \\
    Qwen-7B & MiKV (BS=40) & 2-2 & 0.00 & 0.00 & 0.00 & 0.00 \\ 
    (1$\times$4090-24G) & KIVI (BS=32,40) & 2-2 & 49.29 & 60.31 & 54.57 & 64.57 \\
    \cmidrule(lr){2-7}
    ~ & \methodabbr{} (BS=32) & 2/4-2/4 & 57.35 & 68.03 & \textbf{62.09} & 71.65 \\
    ~ & \methodabbr{} (BS=40) & 2-2 & \textbf{57.58} & \textbf{68.34} & \textbf{62.09} & \textbf{71.81} \\
    
     \midrule
    
    ~ & - - & 16-16 & 57.82 & 68.98 & 61.61 & 71.81 \\
    ~ & RotateKV (BS=6,8) & 4-4 & 58.06 & 68.50 & 61.37 & 71.50 \\
    DeepSeek- & MiKV (BS=6) & 4/16-4/16 & 44.08 & 56.38 & 46.92 & 59.53 \\
    LLaMA-8B & MiKV (BS=8) & 4-4 & 55.69 & 66.61 & 59.95 & 70.39 \\
    (1$\times$4090-24G) & KIVI (BS=6,8) & 4-4 & 57.35 & 68.35 & 71.14 & 71.81 \\
    \cmidrule(lr){2-7}
    ~ & \methodabbr{} (BS=6) & 4/8-4/8 & \textbf{58.77} & \textbf{69.61} & \textbf{63.74} & \textbf{73.70} \\
    ~ & \methodabbr{} (BS=8) & 4-4 & 57.58 & 68.19 & 61.14 & 71.34 \\
    
    \midrule
    
    \multirow{4}{*}{\makecell[l]{DeepSeek-\\Qwen-14B\\(1$\times$A100-40G)}} & - - & 16-16 & 70.14 & 78.74 & 74.40 & 81.73 \\
    ~ & KIVI (BS=12,16) & 2-2 & 67.54 & 77.01 & 72.04 & 80.47 \\
    \cmidrule(lr){2-7}
    ~ & \methodabbr{} (BS=12) & 2/4-2/4 & \textbf{73.70} & \textbf{80.47} & \textbf{77.73} & \textbf{83.46} \\
    ~ & \methodabbr{} (BS=16) & 2-2 & 73.46 & \textbf{80.47} & 77.49 & \textbf{83.46} \\
    
    \midrule
    
    \multirow{4}{*}{\makecell[l]{DeepSeek-\\Qwen-32B\\(1$\times$A100-80G)}} & - - & 16-16 & 74.41 & 81.73 & 78.20 & 84.41 \\
    ~ & KIVI (BS=12,16) & 2-2 & 72.51 & 79.84 & 76.07 & 82.36 \\
    \cmidrule(lr){2-7}
    ~ & \methodabbr{} (BS=12) & 2/4-2/4 & 75.83 & 83.15 & \textbf{78.91} & \textbf{85.35} \\
    ~ & \methodabbr{} (BS=16) & 2-2 & \textbf{76.07} & \textbf{83.30} & \textbf{78.91} & \textbf{85.35} \\

    \midrule
    
    ~ & - - & 16-16 & 82.94 & 88.03 & 86.97 & 90.71 \\
    QwQ-32B & KIVI (BS=12,16) & 2-2 & 73.22 & 80.00 & 78.67 & 84.09 \\
    \cmidrule(lr){2-7}
    (1$\times$A100-80G) & \methodabbr{} (BS=12) & 2/4-2/4 & \textbf{81.99} & \textbf{86.77} & \textbf{85.55} & \textbf{89.45} \\
    ~ & \methodabbr{} (BS=16) & 2-2 & 81.75 & 86.61 & \textbf{85.55} & \textbf{89.45} \\
    \bottomrule
  \end{tabular}
}
\end{table}

\subsection{Comparison with KIVI of Different Bit-widths}
\label{add:kivi}

In our experiments in \cref{sec:exp-main}, \methodabbr{} occupies more memory than KIVI before the KV cache bit-width is reduced to the final Fbit. However, under uniform Fbit settings, both our approach and KIVI consume the same amount of memory once the KV cache bit-width is reduced to Fbit. In fact, under the same GPU and batch size constraints, if KIVI adopts a higher quantization bit-width, it will consume more memory and therefore exhaust the available memory budget before reaching the maximum output length. We evaluate KIVI across different bit-width settings. When KIVI fully utilizes the memory budget, we truncate its output accordingly. As shown in \cref{tab:add-kivi}, our method still outperforms KIVI by 0.83\%–18.33\%.

\begin{table}
  \caption{Results of long-CoT Language Models on reasoning-related benchmarks with KIVI of different bit-widths. ``BS'' is short for ``batch size''.}
  \label{tab:add-kivi}
  \centering
  % \resizebox{\linewidth}{!}{
  \begin{tabular}{llcccccc}
    \toprule
    Models & Quantization & Bit-width & \multicolumn{2}{c}{AIME-2024} & \multicolumn{2}{c}{AIME-2025} \\
    \cmidrule(lr){3-3}\cmidrule(lr){4-5}\cmidrule(lr){6-7}
    (Target GPU) & Methods & (K-V) & pass@1 & Voting & pass@1 & Voting \\

    \midrule
    
    ~ & - - & 16-16 & 41.04 & 63.33 & 30.00 & 36.67 \\
    ~ & KIVI (BS=32,40) & 16-16 & 21.88 & 36.67 & 21.67 & 26.67 \\
    DeepSeek- & KIVI (BS=32,40) & 8-8 & 36.67 & 63.33 & 24.59 & 33.33 \\
    Qwen-7B & KIVI (BS=32,40) & 4-4 & 39.38 & 63.33 & 26.46 & 36.67 \\
    (1$\times$4090-24G) & KIVI (BS=32,40) & 2-2 & 32.08 & 43.33 & 24.58 & 33.33 \\
    \cmidrule(lr){2-7}
    ~ & \methodabbr{} (BS=32) & 2/4-2/4 & \textbf{40.21} & \textbf{66.67} & \textbf{28.96} & \textbf{40.00} \\
    ~ & \methodabbr{} (BS=40) & 2-2 & 40.00 & 60.00 & 28.12 & 33.33 \\

    \bottomrule
  \end{tabular}
  % }
\end{table}

\subsection{Comparison with Mixed-precision Quantization Baselines}
\label{add:kvtuner}

We conduct additional experiments comparing our method with KVTuner on DeepSeek-R1-Distill-Qwen-7B under the same target GPU and batch size settings. For KVTuner, we adopt per-channel asymmetric quantization for the Key cache and per-token asymmetric quantization for the Value cache.  As shown in \cref{tab:add-mixed}, the pass@1 of our method surpasses KVTuner by 2.71-6.04\%.

\begin{table}
  \caption{Results of long-CoT Language Models on reasoning-related benchmarks with the mixed-precision quantization baseline. ``BS'' is short for ``batch size''.}
  \label{tab:add-mixed}
  \centering
  % \resizebox{\linewidth}{!}{
  \begin{tabular}{llcccccc}
    \toprule
    Models & Quantization & Bit-width & \multicolumn{2}{c}{AIME-2024} & \multicolumn{2}{c}{AIME-2025} \\
    \cmidrule(lr){3-3}\cmidrule(lr){4-5}\cmidrule(lr){6-7}
    (Target GPU) & Methods & (K-V) & pass@1 & Voting & pass@1 & Voting \\

    \midrule
    
    ~ & - - & 16-16 & 41.04 & 63.33 & 30.00 & 36.67 \\
    DeepSeek- & KIVI (BS=32) & 2-2 & 32.08 & 43.33 & 24.58 & 33.33 \\
    Qwen-7B & KVTuner (BS=32) & 2/4/8-2/4/8 & 34.17 & 56.67 & 26.25 & 33.33 \\
    \cmidrule(lr){2-7}
    (1$\times$4090-24G) & \methodabbr{} (BS=32) & 2/4-2/4 & \textbf{40.21} & \textbf{66.67} & \textbf{28.96} & \textbf{40.00} \\
    \bottomrule
  \end{tabular}
  % }
\end{table}

\subsection{Performance across Different Hardware Configurations}
\label{add:hardware}
When the memory capacity of the target GPU increases, \methodabbr{} can either maintain the per-request memory allocation by increasing the batch size or allocate a higher Fbit for each layer, allowing the KV cache to use higher bit-widths. We evaluate our method across different GPU memory capacities by increasing the Fbit accordingly. As shown in \cref{tab:add-hardware}, our method consistently achieves accuracy comparable to the original LLM under varying Fbit settings.

\begin{table}[h]
  \caption{Performance of \methodabbr{} across different target GPUs. ``BS'' is short for ``batch size''.}
  \label{tab:add-hardware}
  \centering
  % \resizebox{\linewidth}{!}{
  \begin{tabular}{lcccccc}
    \toprule
    \multirow{2}{*}{Models} & \multirow{2}{*}{Target GPU} & Quantization & Bit-width & \multicolumn{2}{c}{AIME-2024}\\
    \cmidrule(lr){4-4}\cmidrule(lr){5-6}
    & & Methods & (K-V) & pass@1 & Voting \\

    \midrule
    
    % ~ & - - & - - & 16-16 & 41.04 & 63.33 \\
    ~ & \multirow{2}{*}{1$\times$4090-24G} & \methodabbr{} (BS=32) & 2/4-2/4 & 40.21 & 66.67 \\
     & ~ & \methodabbr{} (BS=40) & 2-2 & 40.00 & 60.00 \\
    \cmidrule(lr){2-6}
    DeepSeek- & \multirow{2}{*}{1$\times$A100-40G} & \methodabbr{} (BS=36) & 4/8-4/8 & 40.63 & 63.33 \\
    Qwen-7B & ~ & \methodabbr{} (BS=58) & 4-4 & 40.83 & 66.67 \\
    \cmidrule(lr){2-6}
    & \multirow{2}{*}{1$\times$A100-80G} & \methodabbr{} (BS=108) & 8/16-8/16 & 42.50 & 66.67 \\
    ~ & ~ & \methodabbr{} (BS=164) & 8-8 & 41.87 & 63.33 \\

    \bottomrule
  \end{tabular}
  % }
\end{table}

\subsection{Integration with Sparse Attention}
The progressive quantization and the block-wise memory allocation can be applied naturally to sparse attention mechanisms.
Calibration with positional interpolation is compatible with sparse attention mechanisms that rely on RoPE.
We combine our method with QuestAttention~\cite{quest} and evaluate the accuracy on DeepSeek-R1-Distill-Qwen-7B model. 
As shown in \cref{tab:add-sparse}, our method does not degrade the pass@1 performance of QuestAttention.

\begin{table}[h]
  \caption{Results of \methodabbr{} combined with QuestAttention. ``BS'' is short for ``batch size''.}
  \label{tab:add-sparse}
  \centering
  % \resizebox{\linewidth}{!}{
  \begin{tabular}{llccc}
    \toprule
    Models & \multirow{2}{*}{Methods} & \multicolumn{2}{c}{AIME-2024} \\
    \cmidrule(lr){3-4}
    (Target GPU) &  & pass@1 & Voting \\

    \midrule
    
    \multirow{2}{*}{\makecell[l]{DeepSeek-\\Qwen-7B\\}} & - - & 41.04 & 63.33 \\
     & QuestAttention & 33.33 & 63.33 \\
    (1$\times$4090-24G) & QuestAttention+\methodabbr{} & 33.58 & 60.00 \\
    \bottomrule
  \end{tabular}
  % }
\end{table}

\subsection{Efficiency Analysis of Pre-inference Process}
\label{sec:pre-inference-time}
Before the inference process, \methodabbr{} performs block-wise memory allocation and calibration with positional interpolation as preparation.
Following the experimental setup in \cref{sec:exp-setups}, we measure the time required for these pre-inference procedures.
As shown in \cref{tab:pre-inference-time}, compared with QServe~\cite{qserve}, \methodabbr{} leverages positional interpolation to reduce calibration sequence length from 8,192 to 2,048 tokens, substantially reducing the calibration time by up to 77.21\%.
The additional block-wise memory allocation procedure account for 22.50–23.53\% pre-inference time.

\begin{table}[htbp]
    \caption{Latency of block-wise memory allocation and calibration. ``PI'' is short for ``Positional Interpolation''.}
    \centering
    \begin{tabular}{llcccc}
        \toprule
        \multirow{2}{*}{Model} & \multirow{2}{*}{Method} & \multicolumn{2}{c}{Calibration} & \multirow{2}{*}{\makecell[c]{Memory\\Allocation}} & \multirow{2}{*}{Time} \\
        \cmidrule(lr){3-4}
        ~ & ~ & w/o PI & w/ PI & ~ & ~ \\
        
        \midrule
        
        \multirow{3}{*}{\makecell[l]{DeepSeek-\\Qwen-7B\\}} & QServe (search $\alpha$) & \ding{51} &  &  & 52 min \\
        ~ & \methodabbr{} (BS=40) &  & \ding{51} &  & 13 min \\
        ~ & \methodabbr{} (BS=32) &  & \ding{51} & \ding{51} & 17 min \\

        \midrule

        \multirow{3}{*}{\makecell[l]{DeepSeek-\\Qwen-32B\\}} & QServe (search $\alpha$) & \ding{51} &  &  & 187 min \\
        ~ & \methodabbr{} (BS=16) &  & \ding{51} &  & 44 min \\
        ~ & \methodabbr{} (BS=12) &  & \ding{51} & \ding{51} & 57 min \\

        \midrule

        \multirow{3}{*}{\makecell[l]{DeepSeek-\\LLaMA-70B\\}} & QServe (search $\alpha$) & \ding{51} &  &  & 408 min \\
        ~ & \methodabbr{} (BS=16) &  & \ding{51} &  & 93 min \\
        ~ & \methodabbr{} (BS=12) &  & \ding{51} & \ding{51} & 120 min \\
        
        \bottomrule
    \end{tabular}
    \label{tab:pre-inference-time}
\end{table}

\subsection{Analysis of Progressive Quantization}
Long-CoT tasks exhibit highly variable response lengths. For example, \cref{tab:add-length} summarizes the number of responses of DeepSeek-R1-Distill-LLaMA-8B falling into different length ranges.

\begin{table}[h]
    \caption{The number of responses falling into different length ranges.}
    \centering
    \begin{tabular}{cccc}
        \toprule
        Model & Response Length & AIME-2024 & AIME-2025 \\
        
        \midrule
        
         & 0–4K & 40 & 63 \\
        DeepSeek- & 4–8K & 99 & 67 \\
        LLaMA-8B & 8–16K & 138 & 144 \\
         & 16–32K & 203 & 206 \\
        \bottomrule
    \end{tabular}
    \label{tab:add-length}
\end{table}

For shorter responses (e.g., $<$16K tokens in our experiments), our method keeps the KV cache at a higher bit-width than Fbit throughout decoding. This reduces cumulative quantization error compared with using Fbit KV cache, thereby preserving performance. For longer responses (e.g., $>$16K tokens in our experiments), although the KV cache eventually shrinks to Fbit, the first 16K tokens are generated using a higher-precision KV cache. Consequently, the model benefits from more accurate early-stage generations.

\section{Proof of Equivalent Right Shift}
\label{add:proof}
\begin{theorem}[Equivalent Right Shift]
Given a 16-bit floating-point tensor $\mathbf{X}_{\text{BF16}}$, let $\mathbf{X}_{2b}$ and $\mathbf{X}_{b}$ denote the $2b$-bit and $b$-bit quantized tensors of $\mathbf{X}_{\text{BF16}}$, respectively.
Then
\begin{equation}
    \mathbf{X}_{b}=\left((2^{2b}-2^b+1)(\mathbf{X}_{2b}+2^{b-1})\right) >> 3b.
\end{equation}
\end{theorem}

\begin{proof}
Let the zero points be $Z_{2b} = Z_b = Z$. According to \cref{eq:integer_quant_scale}, the scaling factors are given by
\begin{equation}
S_{2b} = \frac{\max(\mathbf{X}_{\mathrm{BF16}}) - Z}{2^{2b} - 1}, \quad S_{b} = \frac{\max(\mathbf{X}_{\mathrm{BF16}}) - Z}{2^{b} - 1}.
\end{equation}
It follows that
\begin{equation}
S_b = (2^b + 1) S_{2b}.
\end{equation}
Define
\begin{equation}
\widetilde{\mathbf{X}}_{2b} = \frac{\mathbf{X}_{\text{BF16}} - Z}{S_{2b}}, \quad \widetilde{\mathbf{X}}_{b} = \frac{\mathbf{X}_{\text{BF16}} - Z}{S_{b}}.
\end{equation}
Then the quantized tensors are obtained by rounding:
\begin{equation}
\mathbf{X}_{2b} = \left\lfloor \widetilde{\mathbf{X}}_{2b} \right\rceil, \quad \mathbf{X}_{b} = \left\lfloor \widetilde{\mathbf{X}}_{b} \right\rceil,
\end{equation}
and we have
\begin{equation}
\widetilde{\mathbf{X}}_{2b} = (2^b + 1) \widetilde{\mathbf{X}}_{b}.
\end{equation}
By the definition of rounding,
\begin{equation}
\mathbf{X}_{2b} - \frac{1}{2} \le \widetilde{\mathbf{X}}_{2b} < \mathbf{X}_{2b} + \frac{1}{2}.
\end{equation}
Dividing both sides by $2^b + 1$ yields
\begin{equation}
\frac{\mathbf{X}_{2b} - \frac{1}{2}}{2^b + 1} \le \widetilde{\mathbf{X}}_{b} < \frac{\mathbf{X}_{2b} + \frac{1}{2}}{2^b + 1}.
\end{equation}
Perform the Euclidean division of $\mathbf{X}_{2b}$ by $2^b + 1$:
\begin{equation}
\mathbf{X}_{2b} = q (2^b + 1) + r, \quad \text{with } 0 \le q\le 2^b-1 \text{, }0 \le r \le 2^b.
\end{equation}
Then,
\begin{equation}
q + \frac{r - \frac{1}{2}}{2^b + 1} \le \widetilde{\mathbf{X}}_{b} < q + \frac{r + \frac{1}{2}}{2^b + 1}.
\end{equation}
Now consider the expression:
\begin{equation}
\begin{aligned}
\left( (2^{2b} - 2^b + 1)(\mathbf{X}_{2b} + 2^{b-1}) \right) >> 3b 
&= \left\lfloor \frac{(2^{2b} - 2^b + 1)(q(2^b + 1) + r + 2^{b-1})}{2^{3b}} \right\rfloor \\
&= q + \left\lfloor \frac{q + (2^{2b} - 2^b + 1)(r + 2^{b-1})}{2^{3b}} \right\rfloor.
\end{aligned}
\end{equation}
We proceed by considering two cases for the remainder $r$:

\noindent \textbf{Case 1:} $0 \le r \le 2^{b-1}$.

Then,
\begin{equation}
q - \frac{1}{2} < q - \frac{\frac{1}{2}}{2^b + 1} \le \widetilde{\mathbf{X}}_{b} < q + \frac{2^{b-1} + \frac{1}{2}}{2^b + 1} = q + \frac{1}{2}.
\end{equation}
Hence, rounding gives $\mathbf{X}_b = \left\lfloor \widetilde{\mathbf{X}}_b \right\rceil = q$.

Moreover,
\begin{equation}
\frac{q + (2^{2b} - 2^b + 1)(r + 2^{b-1})}{2^{3b}} \ge \frac{(2^{2b} - 2^b + 1) \cdot 2^{b-1}}{2^{3b}} > \frac{2^{2b} \cdot 2^{b-1}}{2^{3b}} = \frac{1}{2} > 0,
\end{equation}
and
\begin{equation}
\begin{aligned}
\frac{q + (2^{2b} - 2^b + 1)(r + 2^{b-1})}{2^{3b}} &\le \frac{2^b - 1 + (2^{2b} - 2^b + 1)(2^{b-1} + 2^{b-1})}{2^{3b}} \\
&= 1-\frac{(2^b - 1)^2}{2^{3b}} < 1.
\end{aligned}
\end{equation}
Therefore,
\begin{equation}
\left\lfloor \frac{q + (2^{2b} - 2^b + 1)(r + 2^{b-1})}{2^{3b}} \right\rfloor = 0,
\end{equation}
and thus
\begin{equation}
\left( (2^{2b} - 2^b + 1)(\mathbf{X}_{2b} + 2^{b-1}) \right) >> 3b = q = \mathbf{X}_{b}.
\end{equation}

\noindent \textbf{Case 2:} $2^{b-1} + 1 \le r \le 2^b$.

Then,
\begin{equation}
q + \frac{1}{2} = q + \frac{2^{b-1} + 1 - \frac{1}{2}}{2^b + 1} \le \widetilde{\mathbf{X}}_{b} < q + \frac{2^b + \frac{1}{2}}{2^b + 1} < q + 1.
\end{equation}
Thus, rounding gives $\mathbf{X}_b = \left\lfloor \widetilde{\mathbf{X}}_b \right\rceil = q + 1$.

Moreover,
\begin{equation}
\frac{q + (2^{2b} - 2^b + 1)(r + 2^{b-1})}{2^{3b}} \ge \frac{(2^{2b} - 2^b + 1)(2^{b-1} + 1 + 2^{b-1})}{2^{3b}} = \frac{2^{3b} + 1}{2^{3b}} > 1,
\end{equation}
and
\begin{equation}
\begin{aligned}
\frac{q + (2^{2b} - 2^b + 1)(r + 2^{b-1})}{2^{3b}} &\le \frac{2^b - 1 + (2^{2b} - 2^b + 1)(2^b + 2^{b-1})}{2^{3b}} \\
&= \frac{2^{3b} + 2^{3b-1} - 2^{2b} - 2^{2b-1} + 2^{b+1} + 2^{b-1} - 1}{2^{3b}} \\
&= 2 - \frac{2^{3b-1} + 2^{2b} + 2^{2b-1} - 2^{b+1} - 2^{b-1} + 1}{2^{3b}} < 2.
\end{aligned}
\end{equation}
Therefore,
\begin{equation}
\left\lfloor \frac{q + (2^{2b} - 2^b + 1)(r + 2^{b-1})}{2^{3b}} \right\rfloor = 1,
\end{equation}
and thus
\begin{equation}
\left( (2^{2b} - 2^b + 1)(\mathbf{X}_{2b} + 2^{b-1}) \right) >> 3b = q + 1 = \mathbf{X}_{b}.
\end{equation}

\noindent In both cases, the desired equality holds, which completes the proof.
\end{proof}

\section{Limitations} 
In this paper, we do not consider all of the attention mechanisms, such as the multi-head latent attention (MLA), which is quite different from the widely used Group-Query Attention (GQA).
Besides, we do not combine the proposed \methodabbr{} with other system-level optimization techniques and inference engines, which yields for future work.

% \section{Ethics Statement}
% This work focuses on reducing the substantial overhead caused by the linearly growing KV cache in long-context processing through KV Cache quantization.
% On the one hand, the proposed \methodabbr{} better preserves model accuracy after low-precision KV cache quantization, making it more accessible for cost-constrained institutions, individuals, and application scenarios.
% On the other hand, as a lossy compression technique, quantization can introduce distribution shifts and performance degradation, potentially leading to increased hallucinations or instruction-following failures.
% Therefore, additional caution and oversight are required during deployment.

% \section{Appendix}
% You may include other additional sections here.

\end{document}